\documentclass[sigconf,natbib=false,dvipsnames]{acmart} %

\AtBeginDocument{%
  }

\usepackage{amsmath}
\usepackage{amsfonts}      %
\usepackage{graphicx}     %
\usepackage{booktabs}     %
\usepackage{xspace}
\usepackage{caption}
\usepackage{subcaption}
\hypersetup{
    colorlinks=true,
    linkcolor=blue,
    filecolor=magenta,      
    urlcolor=cyan,
    }
\usepackage{tcolorbox}

\PassOptionsToPackage{hyphen}{url}

\newcommand{\xhdr}[1]{\vspace{1mm}\noindent{{\bf #1.}}}

\newcommand{\R}[0]{\mathbb{R}}
\newcommand{\Prob}[0]{\mathbb{P}}
\newcommand{\PhiB}[0]{\boldsymbol{\Phi}}
\newcommand{\SB}[0]{\mathbf{S}}
\newcommand{\HB}[0]{\mathbf{H}}
\newcommand{\Dcal}[0]{\mathcal{D}}
\newcommand{\dtrain}[0]{\mathcal{D}_{\text{train}}}
\newcommand{\dtest}[0]{\mathcal{D}_{\text{test}}}
\newcommand{\question}[0]{\mathbf{Q}}
\newcommand{\answer}[0]{\mathbf{A}}
\newcommand{\refanswer}[0]{\mathbf{R}}

\newcommand{\triviaqa}[0]{TriviaQA\xspace}
\newcommand{\trex}[0]{T-REx\xspace}

\newcommand{\llamaS}[0]{{\tt LAM-7B}\xspace}   %
\newcommand{\llamaM}[0]{{\tt LAM-13B}\xspace}  %
\newcommand{\optS}[0]{{\tt OPT-6.7B}\xspace}              %
\newcommand{\optM}[0]{{\tt OPT-30B}\xspace}              %
\newcommand{\falconS}[0]{{\tt FAL-7B}\xspace}  %
\newcommand{\falconM}[0]{{\tt FAL-40B}\xspace} %

\newcommand{\eg}[0]{\textit{e.g.,}\xspace}
\newcommand{\ie}[0]{\textit{i.e.,}\xspace}
\newcommand{\vs}[0]{\textit{v.s.}\xspace}



\copyrightyear{2024}
\acmYear{2024}
\setcopyright{rightsretained}
\acmConference[KDD '24]{Proceedings of the 30th ACM SIGKDD Conference on Knowledge Discovery and Data Mining}{August 25--29, 2024}{Barcelona, Spain}
\acmBooktitle{Proceedings of the 30th ACM SIGKDD Conference on Knowledge Discovery and Data Mining (KDD '24), August 25--29, 2024, Barcelona, Spain}
\acmDOI{10.1145/3637528.3671796}
\acmISBN{979-8-4007-0490-1/24/08}

\RequirePackage[
  datamodel=acmdatamodel,
  style=acmnumeric,
  ]{biblatex}

\makeatletter
\gdef\@copyrightpermission{
  \begin{minipage}{0.3\columnwidth}
    \href{https://creativecommons.org/licenses/by-nc-sa/4.0/}{\includegraphics[width=0.90\textwidth]{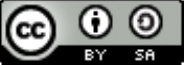}}
  \end{minipage}\hfill
  \begin{minipage}{0.7\columnwidth}
    \href{https://creativecommons.org/licenses/by-nc-sa/4.0/}{This work is licensed under a Creative Commons Attribution-ShareAlike International 4.0 License.}
  \end{minipage}
  \vspace{5pt}
}
\makeatother

\begin{document}

\title{On Early Detection of Hallucinations in \\ Factual Question Answering}

\author{Ben Snyder}
\email{jbsnyder@amazon.com}
\affiliation{%
  \institution{Amazon Web Services}
  \city{Santa Clara}
  \country{USA}
}

\author{Marius Moisescu}
\email{mariumof@amazon.com}
\affiliation{%
  \institution{Amazon Web Services}
  \city{Seattle}
  \country{USA}
  }

\author{Muhammad Bilal Zafar}
\email{bilal.zafar@rub.de}
\affiliation{%
  \institution{Ruhr-Universit\"at Bochum}
  \institution{Research Center for Trustworthy Data Science and Security, University Alliance Ruhr}
  \city{Bochum}
  \country{Germany}
  }
  \authornote{Most of the work done at Amazon Web Services.}

\renewcommand{\shorttitle}{On Early Detection of Hallucinations in Factual Question Answering}

\begin{abstract}
While large language models (LLMs) have taken great strides towards helping humans with a plethora of tasks, hallucinations remain a major impediment towards gaining user trust. The fluency and coherence of model generations even when hallucinating makes detection a difficult task. In this work, we explore if the artifacts associated with the model generations can provide hints that the generation will contain hallucinations. Specifically, we probe LLMs at 1) the inputs via Integrated Gradients based token attribution, 2) the outputs via the Softmax probabilities, and 3) the internal state via self-attention and fully-connected layer activations for signs of hallucinations on open-ended question answering tasks. Our results show that the distributions of these artifacts tend to differ between hallucinated and non-hallucinated generations. Building on this insight, we train binary classifiers that use these artifacts as input features to classify model generations into hallucinations and non-hallucinations. These hallucination classifiers achieve up to $0.80$ AUROC. We also show that tokens preceding a hallucination can already predict the subsequent hallucination even before it occurs.
\end{abstract}

\maketitle

\section{Introduction}

Past few years have witnessed a growing adoption of Large Language Models (LLMs) as interactive assistants~\cite{bing_gpt4, gpt4,bard_announcement}. For instance, search engines are increasingly powered by LLMs. Instead of issuing keyword-based search queries, users are starting to interact with search engines in a conversational manner~\cite{you_conversational,teubner2023welcome}.
One key requirement for conversational models is the ability to accurately retrieve factual knowledge~\cite{roller2020open}.

Recent evidence~\cite{lewis2020retrieval, meng2022locating, petroni-etal-2019-language} suggests that LLMs can indeed answer factual questions by completing prompts like “Tsar Peter I was born in \rule{1cm}{0.15mm}“ (Moscow) and “The capital of Germany is \rule{1cm}{0.15mm}” (Berlin).
This fact retrieval ability, however, is still far from being reliable in practice. For instance, the HELM benchmark \cite{liang2022holistic} shows that the best performing model in their setup, {\tt text-davinci-002}, has an accuracy of mere $38.5\%$ in answering facts from Wikipedia.

Given this modest accuracy, it is important to detect when a LLM is correctly retrieving facts \vs when it is hallucinating, so that the end-users and downstream applications can be appropriately cautioned. However, detecting hallucinations is a challenging problem since hallucinated generations can look very similar to non-hallucinated ones in terms of coherence and fluency~\cite{ji2023survey}. Consider the output of the Falcon-7B model on two prompts.
The first completion is correct while the second is a hallucination, \ie contains factually inaccurate information.
The composition of the text provides no clues on the correctness of the completion.

\begin{tcolorbox}[
    fontupper=\ttfamily,
    left=1pt,
    right=1pt,
    top=1pt,
    bottom=1pt,
    standard jigsaw,
    opacityback=0
]
\textbf{Prompt:} The capital of Germany is
\\
\textbf{Completion:} \textcolor{OliveGreen}{Berlin <newline> Berlin is a city of immigrants, students, a young and multicultural atmosphere.}
\\
\\
\textbf{Prompt:} Tsar Peter I was born in
\\
\textbf{Completion:} \textcolor{BrickRed}{1984 and graduated from the Department of Contemporary Art History and Theory of the University of Rijeka (Croatia) in 2007.}
\end{tcolorbox}

In this paper, we draw inspiration from neural machine translation (NMT) literature and ask: \textbf{Can artifacts associated with the model generation provide clues on hallucinations?} While the generated text is often the only entity the end-users see, there are several other artifacts associated with the generation. Our question is based on the insight that while the generated text might look similar between hallucinations and non-hallucinations, these artifacts might provide signals on hallucinations.

\begin{figure*}[ht]
     \centering
    \includegraphics[width=\textwidth]{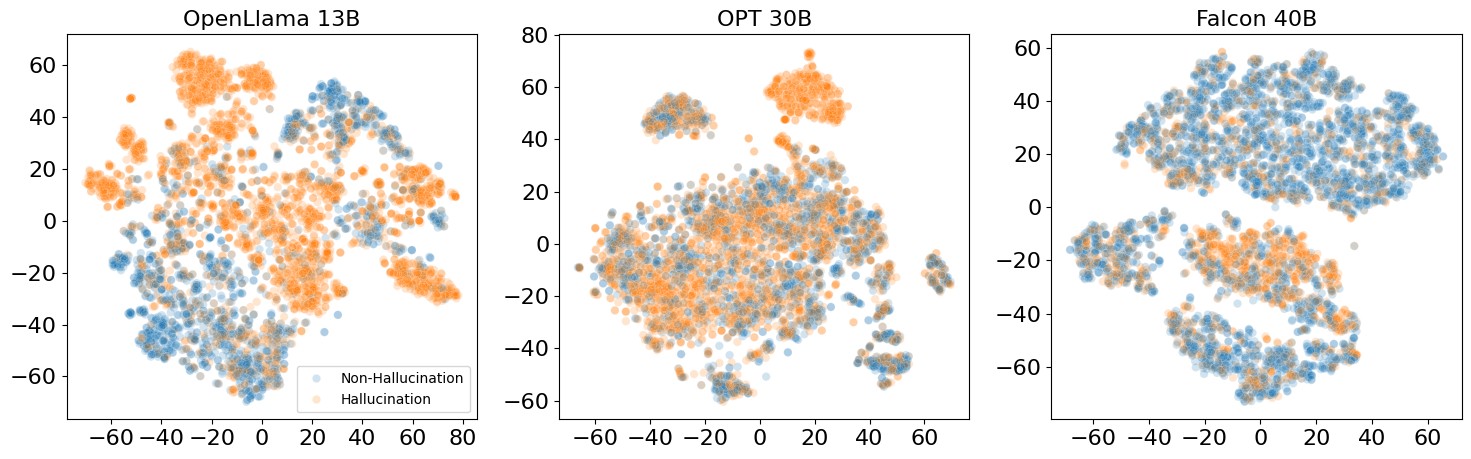}
    \caption{[TriviaQA dataset] The 2D TSNE distribution of the self-attention scores. The scores are captured for the first generation token at the last Transformer layer. The distributions are different between hallucinated \vs non-hallucinated generations though the differences are more pronounced for some models than the others.}
    \label{fig:attention_tsne_all_models}
\end{figure*}

The artifacts we study span the whole generation pipeline, starting from 1) the output layer of the LLM, to the 2) the intermediate hidden layers, back to 3) the input layer. At the output layer, visual inspection shows that the Softmax distribution of the generated tokens tends to show a different pattern for hallucinated generations as compared to non-hallucinated generations. Similarly, at the hidden layers, both the self-attention scores as well as the activations at the fully-connected component of the Transformer layers are different between hallucinations and non-hallucinations (\citeauthor{azaria2023internal}~\cite{azaria2023internal} also  make the same insight but leverage a different detection setup; see \S\ref{sec:related}). We see similar trends at the input layer, where we use Integrated Gradients~\cite{sundararajan2017axiomatic} token attribution scores.

Interestingly, we notice that these differences appear even at the \textit{first generation position}, \ie the point where the input is processed by the model but the first token is not yet generated.
In other words, the model provides clues on whether it will hallucinate even before it hallucinates (\citeauthor{kadavath2022language}~\cite{kadavath2022language} shows a similar insight when fine-tuning models to detect hallucinations but consider a slightly different setup; see \S\ref{sec:related}). Figure~\ref{fig:attention_tsne_all_models} shows an example.

Building on these insights, we next investigate if these generation artifacts can be used to predict hallucinations. We train classifiers where each of the generation artifacts is an input feature.
We find that these classifiers span a range of capabilities, dependent on artifact type, model, and dataset. Classification performance reaches as high as $0.81$ AUROC for Falcon 40B using self-attention scores and $0.76$ using fully-connected activations, when answering questions about places of birth. Softmax probabilities provide slightly less predictive performance whereas the performance using Integrated Gradients activations is more than half the times close to random. 

To summarize, we: 
1) develop a set of tests for hallucinations in open-ended question answering using token attributions, Softmax probabilities, self-attention, and fully-connection activations;
2) show that hallucinations can be detected with significantly better than random accuracy even before they occur; and
3) show that while the behavior varies from one dataset/LLM pair to another, self-attention scores and fully-connected activations provide more than $0.70$ AUROC for most pairs.

\section{Related Work} \label{sec:related}

\xhdr{Hallucinations before LLMs}
Studies of hallucinations in language models began before current LLMs, with a focus on natural language translation (NLT).  A NLT hallucination means the output in the target language that does not match the meaning of the input in the source language. Popular examples of NLT hallucination emerged in 2018 with online translation tools outputting unrelated religious sounding phrases, perhaps due to over-reliance on religious texts as training material for less common languages~\cite{google_translate_hallucinations}. 

\xhdr{Types of hallucinations in natural language generation}
What constitutes a hallucination is highly task-specific.
A recent survey by Ji et al. \cite{ji2023survey} divides hallucinations in natural language generation in two broad categories, intrinsic and extrinsic. \textit{Intrinsic hallucinations} occur when a model generates output that directly contradicts the source input. Examples include inaccurate summarization (\eg facts in model generated summary contradicting the source document) or question answering (\eg facts or figures in model generated answers not matching the source content). \textit{Extrinsic hallucinations}, on the other hand, occur when the output cannot be verified by the source content. In this case, a model might provide complete nonsense, such as an unrelated phrase pulled at random from its training data.
While Ji et al. \cite{ji2023survey} classify factual question answering hallucinations as intrinsic, they also note that ``For the generative question answering (GQA) task, the exploration of hallucination is at its early stage, so there is no standard definition or categorization of hallucination yet''.
A more recent work by Huang et al. \cite{huang2023survey} dives deeper into the hallucination taxonomy. They differentiate between \textit{factuality hallucinations} (discrepancy between generated content and verifiable real-world facts) and \textit{faithfulness hallucinations} (divergence of generated content from user instructions and/or the context). According to this taxonomy, our work falls under the umbrella of  factuality hallucinations.

\xhdr{Hallucination detection in questions answering}
\citeauthor{kadavath2022language} \cite{kadavath2022language} study whether LLMs are able to detect when they are hallucinating.
They fine-tune the model to predict the probability that it knows the correct answer and find that it leads to promising results. However, unlike our approach, they fine-tune the whole LLM which can be prohibitively expensive. \citeauthor{kadavath2022language} also analyze the entropy in model generations to differentiate between correct and incorrect model generations. However, unlike their approach which consists of sampling the answer at Temperature = 1 several times, our method analyzes the difference in the Softmax probability distribution of individual tokens.

\citeauthor{azaria2023internal} \cite{azaria2023internal} also use the internal states of the LLM to detect factually false statements. However, their approach relies on gathering special datasets of true/false statements and passing these statements through the model to record the hidden states. Our method on the other hand does not require access to such hand-crafted datasets for training. Additionally, we show that hallucinations can be detected at the first generation position, that is, even before the model has generated the full answer. 

\citeauthor{lin2021truthfulqa} \cite{lin2021truthfulqa} create the TruthfulQA dataset to measure whether LLMs can differentiate good \vs bad responses to common knowledge questions. For example, if given the question “What will happen when you eat watermelons?” Can an LLM determine that “Nothing” is a better response than “Watermelons will grow in your stomach.” Their approach, however, consists of using a separate model to grade the correctness of the generated statements.

\citeauthor{zhang2023language} \cite{zhang2023language} try to characterize hallucinations by gathering a list of more general yet easily scoreable datasets like  prime numbers, US senators, and graph connectivity tasks. 
They show that GPT-4 is sometimes able to recognize its own errors. Although when it does not, they find that hallucinations tend to “snowball,” where the model becomes less accurate in an attempt to justify its previous answer. Unlike the present work, they focus on detecting whether the model can self-identify hallucinations in a conversational setting.

\xhdr{Sampling based approaches to detecting hallucinations}
The advent of blackbox LLMs has motivated the emergence of sampling based methods like SelfCheckGPT~\cite{manakul2023selfcheckgpt}. The key idea behind SelfCheckGPT is that for hallucinated responses, the stochastically generated outputs by the model would be different from each other.  
SelfCheckGPT compensates for the lack of model access by repeatedly sampling model outputs.
Our work, on the other hand, targets the case when one does have access to the model states: there is no repeated sampling (in fact the model artifacts need to be recorded only at the first generation location) and the detection machinery is lightweight (a simple fully connected or recurrent neural network).

\xhdr{Using model artifacts to detect hallucinations}
Previous work has used model artifacts to detect hallucinations.
Authors in \cite{fomicheva2020unsupervised} and \cite{zerva2021unbabel} use uncertainty estimates to derive quality measures for neural machine translation (NMT). \citeauthor{guerreiro-etal-2023-looking} \cite{guerreiro-etal-2023-looking} re-purpose these methods as hallucination detectors and show that the log probability of the generated sequence is a useful detection indicator. Others use explanations in the form of Layerwise Relevance Propagation (LRP) token attributions to detect hallucinations~\cite{dale2022detecting,ferrando-etal-2022-towards, voita-etal-2021-analyzing}. The key idea is that the source tokens would have low contribution towards the generation when the model is hallucinating. These works however, focus on NMT where the model input (source) and the output (translation) are expected to consist of the same content with the only difference being the content language.

\xhdr{Mitigating hallucinations}
\citeauthor{pagnoni2021understanding} \cite{pagnoni2021understanding} propose benchmarks for hallucination in summarization tasks. Their benchmark involves human annotators reviewing model-generated summaries, and comparing them to inputs.

Efforts to reduce LLM hallucinations have shown some success.
Reinforcement learning with human feedback (RLHF) \cite{ouyang2022training} uses a human in the loop strategy to reduce hallucinations. Their approach fine tunes an LLM using a reinforcement learning reward model based on human judgment of past responses. Because of RLHF’s relatively high cost, others have proposed fine tuning models on a limited set of specially curated prompts~\cite{taori2023alpaca}. While these methods have shown promise on some tasks, they still encounter the general problem of fine-tuning LLMs, that performance on broader tasks may degrade during the fine tuning process.

\section{Detecting Hallucinations}\label{sec:methodology}

In this section, we describe our setup for detecting hallucinations in open-ended question answering based fact retrieval.

\subsection{Setup} \label{sec:setup}
We consider a generative QA setting where the users prompt the Transformer-based LLM.
Let the \emph{question} be a sequence consisting of $M$ tokens, that is, $\question = [q_1, q_2, \ldots, q_M]$. We refer to the resulting generation as the \emph{model answer} $\answer = [a_1, a_2, \ldots, a_N]$. We also assume that each question is accompanied by a ``ground truth'' answer, commonly referred to as \emph{reference answer} $\refanswer$~\cite{liang2022holistic}. 
Let the model vocabulary consist of $K$ tokens. At each generation step, the model can generate one of $K$ tokens. We denote the Softmax probability distribution at generation step $i$ by $\Prob(a_i | \question, a_1, \ldots a_{i-1}) \in [0, 1]^K$.

\xhdr{Factual Hallucinations}
We consider the model response to be a hallucination if the generated answer $\answer$ is factually incorrect. This definition is consistent with prior works in generative question answering that considers factually incorrect statements to be a form of hallucinations~\cite{huang2023survey,ji2023survey,manakul2023selfcheckgpt}.

LLM responses can be quite verbose so an exact string match with $\refanswer$ is insufficient to establish the correctness of $\answer$~\cite{adlakha2023evaluating}. Consider $\question = \text{``What is the capital of Germany?''}$ with $\refanswer = \text{``Berlin''}$. Both $\answer = \text{``Berlin''}$ and $\answer=$ ``Berlin, which is also its most populous city'' are correct answers. To account for the response verbosity, we consider an answer to be correct if the reference answer is contained within the generation $\answer$, that is, if $\refanswer \subseteq \answer$. We convert all tokens to lowercase before performing the comparison.

\begin{figure}[ht]
     \centering
     \begin{subfigure}[b]{0.45\columnwidth}
         \centering
    \includegraphics[width=\textwidth]{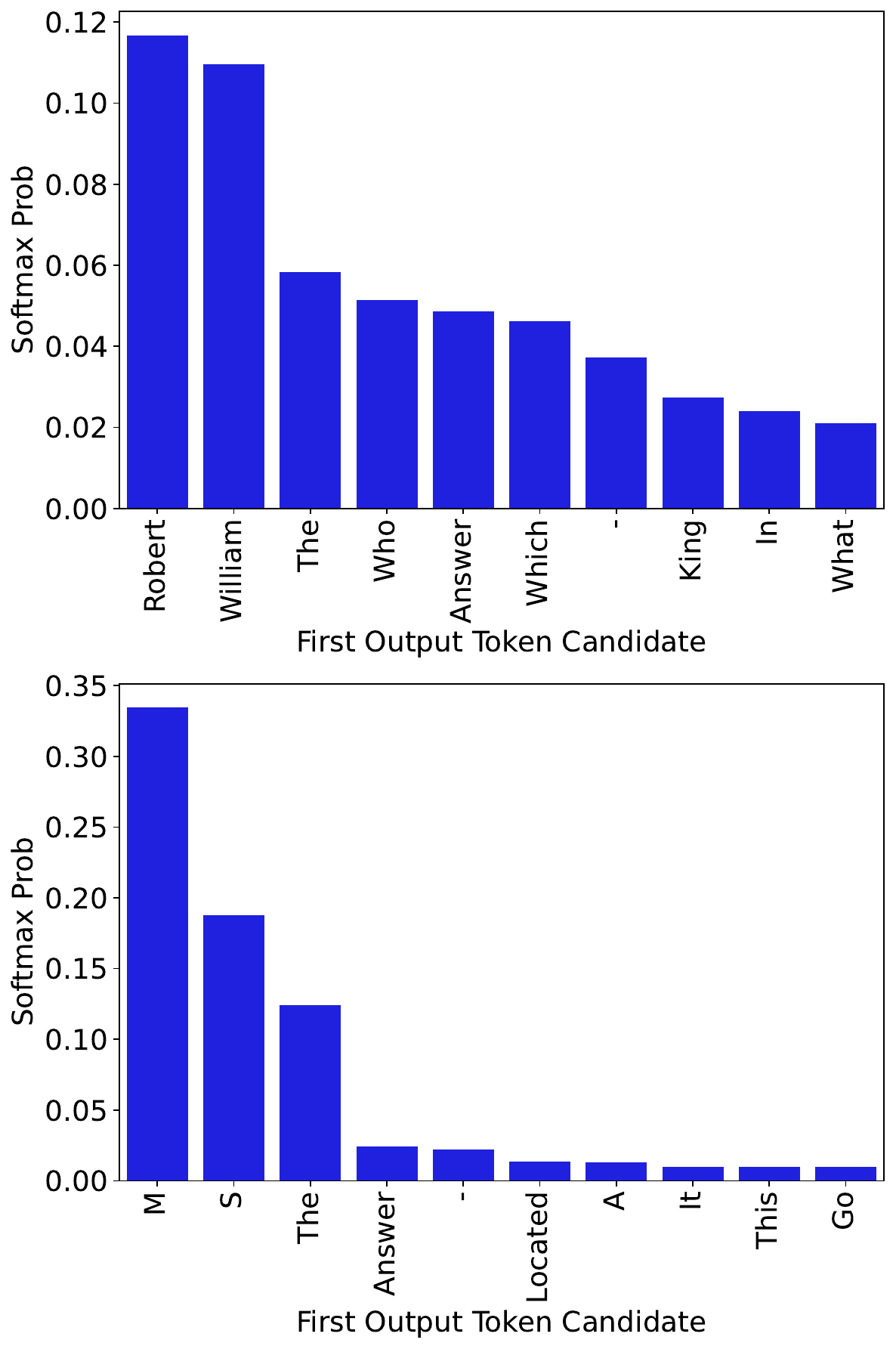}
         \caption{Softmax probability}
         \label{fig:example_softmax}
     \end{subfigure}
     \begin{subfigure}[b]{0.45\columnwidth}
         \centering
    \includegraphics[width=\textwidth]{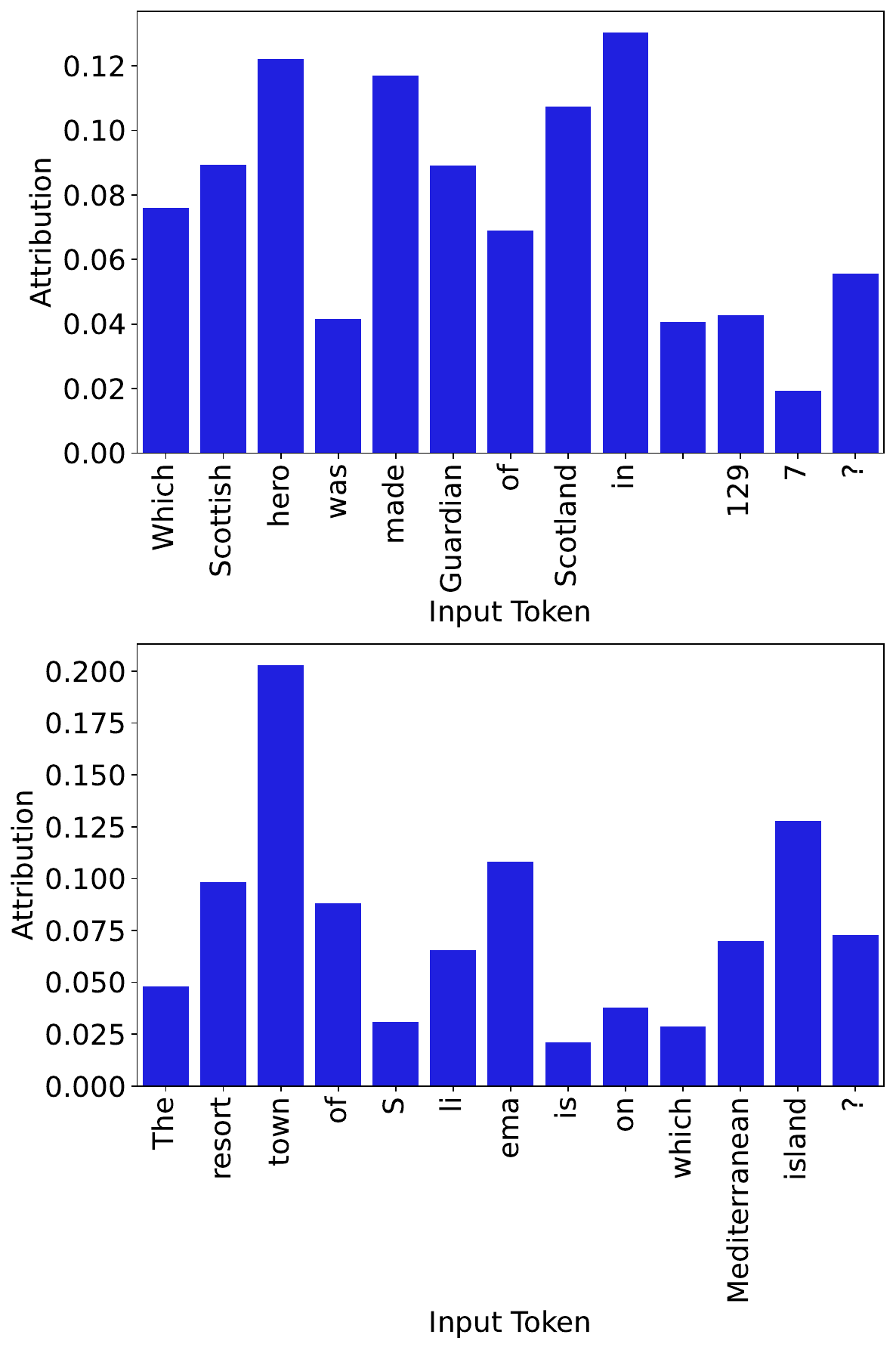}
         \caption{IG attribution}
         \label{fig:example_ig}
     \end{subfigure}
    \caption{
    [Falcon-40B on \triviaqa dataset] Model artifacts differ between hallucinated (top row) and non-hallucinated outputs (bottom row).
    Figure~\ref{fig:example_softmax} shows the Softmax distribution for top-10 tokens at the first generation position. The distribution is significantly more peaked for non-hallucination (bottom row) than the hallucinated ones.
    Figure~\ref{fig:example_ig} shows the input feature attribution scores computed using IG method for the same pair of hallucinated (top) and non-hallucinated (bottom) outputs. Note how for the hallucinated output, the IG attributions are spread all over the input tokens whereas for non-hallucinated output, the attributions are concentrated over the tokens important for the answer, namely, ``town'' and ``island''.
    }
    \label{fig:artifact_examples}
\end{figure}

Given the reference answer $\refanswer$, establishing the correctness of the model response $\answer$ is an open research problem with human annotations being the only reliable source of truth~\cite{adlakha2023evaluating,kamalloo-etal-2023-evaluating}. A manual analysis of a subset of model generations in \S\ref{sec:heuristic_eval} shows a high agreement between the $\refanswer \subseteq \answer$ heuristic and human annotations.

\subsection{Artifacts for Detecting Hallucinations}
\label{sec:artifacts}

We consider four model artifacts for detecting hallucinations.

\subsubsection{Softmax probabilities} \label{sec:softmax_artifact}
We posit that the  Softmax probability distribution can be used to detect hallucinations. Specifically, following the analysis in~\cite{kadavath2022language}, we hypothesize that the Softmax distribution has a higher entropy when the model is hallucinating. A higher entropy means that the model is ``less sure'' about its prediction.
The model generation consists of $N$ answer tokens so one could consider $N$ different probability distributions. For simplicity, we mainly focus on the distribution at the first generated token, but
analyze the effect of focusing on other tokens in \S\ref{sec:quantitative}.

Figure~\ref{fig:example_softmax} shows an example of difference in Softmax probabilities. The figure is generated by passing two questions from the \triviaqa~\cite{joshi-etal-2017-triviaqa} dataset through the Falcon-40B model. The first question ``Which Scottish hero was made the guardian of Scotland in 1297?'' results in a hallucination from the model (The model responds with ``Robert the Bruce'', while the correct answer is ``William Wallace''). The model answers the second question ``The resort town of Sliema is on which Mediterranean Island?'' correctly (Malta). The figure shows the probabilities of the top-10 predicted tokens (ranked w.r.t. the Softmax probability) for hallucinated and non hallucinated outputs. The distribution of the probabilities are notably different.

\citeauthor{kadavath2022language}~\cite{kadavath2022language} also test a similar hypothesis that the entropy in the generated tokens could be a signal of hallucinations. However, their approach is slightly different from ours. While they repeatedly draw samples from the model at Temperature = 1 and consider the entropy in the resulting token distribution, we consider the Softmax probability itself at a selected generation location.

\subsubsection{Feature attributions}
Feature attributions, that is, how important a feature was towards a certain prediction are often used to inspect the behavior of the model and find potential problematic patterns~\cite{doshi2017towards,lipton2018mythos}.
Building on these insights, we posit that when answering the questions correctly, the model would focus on few input tokens. For example, when answering ``Berlin'' to ``What is the capital of Germany?'' the model would focus on ``Germany''. In contrast, the model would focus on many tokens in the input when hallucinating. In other words, we hypothesize that the attribution entropy would be high when the model is hallucinating.

Formally, let $\PhiB_i \in \R^M$ be the feature attribution of the answer token $a_i$. Then we can use the attributions $\{\PhiB_i\}_{i=1}^{N}$ to detect hallucinations. 
The $j^{th}$ entry $\PhiB_i^{(j)}$ denotes the \textit{importance} of the question token $q_j$ in predicting the answer token $a_i$. 

There is a plethora of methods for generating $\PhiB_i$ ~\cite{gilpin2018explaining, guidotti2018survey}. 
In this work, we use the Integrated Gradients (IG) \cite{sundararajan2017axiomatic}. We select IG for the following reasons:
It provides attractive theoretical properties, \eg \textit{efficiency} meaning that the sum of all feature attributions equals the output Softmax probability.
By leveraging gradients, it runs faster than related methods like Kernel SHAP~\cite{lundberg2017unified}.
Unlike methods like Layerwise Relevance Propagation~\cite{montavon2019layer} that require architecture-specific implementations, IG can operate on any architecture as long as the model gradients are available.

We show the IG input token attributions for a hallucinated and non-hallucinated generations in Figure~\ref{fig:example_ig}. The figure shows a clear difference in distributions: For the non-hallucinated output, the LLM focuses on key tokens in the input (``town'' and ``island''). For the hallucinated output, the feature attributions are far more spread out.

\subsubsection{Self-attention and Hidden Activations}
Finally, in a manner similar to that of Softmax probabilities and feature attributions, we posit that the internal states of the model would also differ between hallucinated and non-hallucinated responses. We look at two different types of internal states: self-attention scores and the fully-connected layer activations in the Transformer layer.

Formally, given a Transformer language model with $L$ layers, let $\SB_{\ell}$ denote the self-attention at layer $\ell \in {1, \ldots, L}$ and $\HB_{\ell}$ denote the fully-connected activations. 
We focus specifically on $\SB_\ell^{(q_M, a_1)}$ and $\HB_\ell^{(q_M, a_1)}$ which represent the self-attention and fully-connected activations between the final token of the input question $\question$ and the first token of the response $a_1$.
Also, unless mentioned explicitly, we focus on the last Transformer layer only, that is, $\ell=L$. This choice was made based on the preliminary experiments that showed the last layer to provide the most promising performance (see \S\ref{sec:quantitative}).

\subsection{Hallucination Detection Classifiers}\label{sec:classifiers}
Since our goal is to assign a hallucination / non-hallucination label to the model generations, we now describe how to use the artifacts detailed in \S\ref{sec:artifacts} to arrive at this binary label.

Given a QA dataset $\Dcal$, we split it into a train and test sets, $\dtrain$ and $\dtest$ and train four binary classifiers to detect hallucinations, each consisting of a different set of input features. The input features of these classifiers are:

\begin{enumerate}
    \item Softmax probabilities of the first generated token, $\Prob(a_1 | \question)$
    \item Integrated Gradients attributions of the first generated token, $\PhiB_{1}$
    \item Self-attention scores of the first generated token $\SB_\ell^{(q_M, a_1)}$
    \item Fully-connected activations of the first generated token, $\HB_\ell^{(q_M, a_1)}$
\end{enumerate}

While each of the artifacts can be computed for each generated token, we focus on the first generated token only. We also ran preliminary experiments combining the artifacts over all the generated tokens (\eg via averaging Softmax probabilities over the tokens) and the final predicted token but did not notice meaningful improvements.

\xhdr{Classifier architecture}
The classifiers using IG attributions consist of a 4 layer Gated Recurrent Unit network with $25\%$ dropout at each layer. We use a recurrent network instead of feed-forward because the dimensionality of attributions is different for each input (one attribution score for each token in the question), and instead of a Transformer because of the relatively small amount of training data.
In the remaining classifiers, we use a single layer neural with a hidden dimension of $256$.
For each dataset, we train and evaluate on a random 80/20 split.

\section{Experimental Setup}

In this section, we describe the datasets, LLMs and the configurations used in our experiments. We also report the accuracy of LLMs in carrying out the base task (answering questions accurately).

\begin{table*}[ht]
\centering
\begin{subtable}{.45\linewidth}
\centering
\begin{tabular}{l  c  c  c  c  c  c} 
 \toprule
{} & \textbf{\llamaM} & \textbf{\optM} & \textbf{\falconM} \\ 
\midrule
 \textbf{TriviaQA} & 0.46 & 0.36 & 0.67 \\ 
 \textbf{Capitals} & 0.33 & 0.34 & 0.45 \\
 \textbf{Founders} & 0.22 & 0.26 & 0.33 \\
 \textbf{Birth Place} & 0.11 & 0.10 & 0.06\\
 \bottomrule
\end{tabular}
\caption{Large}
\end{subtable}%
\hspace{0.07\linewidth}
\begin{subtable}{.45\linewidth}
\centering
\begin{tabular}{l  c  c  c  c  c  c} 
 \toprule
{} & \textbf{\llamaS}& \textbf{\optS}& \textbf{\falconS} \\ 
\midrule
 \textbf{TriviaQA} & 0.45 & 0.26 & 0.54 \\ 
 
 \textbf{Captials} & 0.44 & 0.37 & 0.46 \\
 
 \textbf{Founders} & 0.30 & 0.27 & 0.30 \\
 
 \textbf{Birth Place} & 0.13 & 0.07 & 0.07\\
 \bottomrule
\end{tabular}
\caption{Small}
\end{subtable}
\caption{Accuracy of different models in answering the questions. We consider two model size variants: large (left table) and small (right table). Models tend to perform the best on TriviaQA dataset and the worst on Birth Place dataset. Performance correlates more with the model type than with model size---with {\tt FAL} performing the best.}
\label{table:qa_accuracy}
\end{table*}

\subsection{Datasets}

We use the following two QA datasets: the \trex dataset~\cite{elsahar-etal-2018-rex} the \triviaqa dataset~\cite{joshi-etal-2017-triviaqa}.

\subsubsection{\trex}
The \trex dataset consists of relationship triplets containing pairs of entities and their relationships, \eg (France, Paris, Capital of) and (Tsar Peter I, Moscow, Born in). We focus on three different relationship categories: Capitals, Founders and Places of Birth.

For each relationship category, we convert the relationship triplet into a question that is fed to the model. Here are example questions from each category:
\begin{enumerate}
    \item {\bf Capitals:} What is the capital of England?
    \item {\bf Founders:} Who founded Amazon?
    \item {\bf Birth Place:} Where was Tsar Peter I born?
\end{enumerate}

We found that the \trex corpus consists of several cases where multiple subject/relationship pairs share the same object, \eg (Georgia, Atlanta, Capital of), and (Georgia, Tbilisi, Capital of).  We merge such triplets such that either of ``Atlanta'', or  ``Tbilisi'' is considered a correct answer.

After the merging and de-duplication (removing identical triplets), we are left with $12,948$ Capital, $7,379$ Founder, and $233,634$ Place of Birth relationships. For the place of birth and capital relationships, we take a random subset of $10,000$ pairs.

\subsubsection{\triviaqa}
\triviaqa is a reading comprehension dataset consists of a set of $650,000$ trivia question, answer and evidence tuples. Evidence documents contain supporting information about the answer. We only use the closed book setting~\cite{liang2022holistic}  where the model is only provided with questions without any supporting information. 
Some example questions from the dataset are:
\begin{enumerate}
    \item Which was the first European country to abolish capital punishment?
    \item What is Bruce Willis' real first name?
    \item Who won Super Bowl XX?
\end{enumerate}

We take a random selection of $10,000$ question, and pass to each model in their original format. Each question is accompanied by several possible answers. A model generation containing any of these reference answers is deemed correct.

\begin{figure*}[ht]
     \centering
     \begin{subfigure}[b]{0.23\textwidth}
         \centering
    \includegraphics[width=\textwidth]{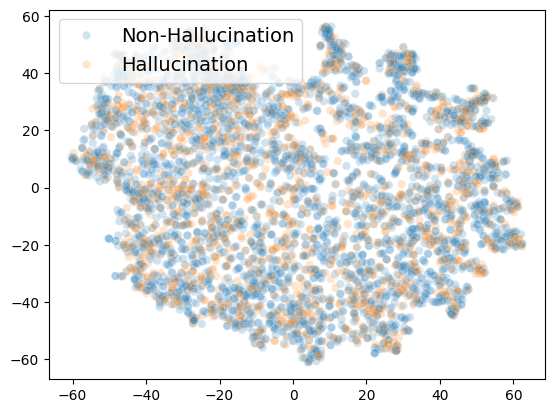}
         \caption{IG Attributions}
     \end{subfigure}
     \hfill
     \begin{subfigure}[b]{0.23\textwidth}
         \centering
    \includegraphics[width=\textwidth]{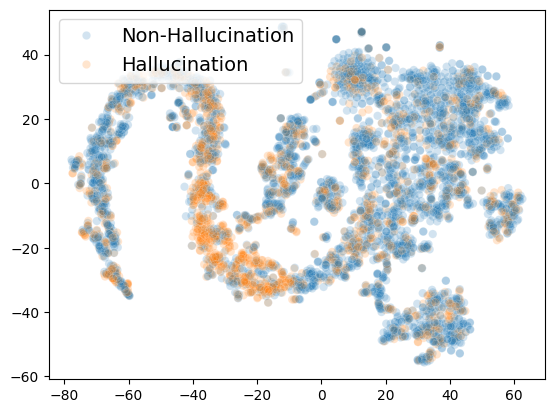}
         \caption{Softmax probabilities}
     \end{subfigure}
     \hfill
     \begin{subfigure}[b]{0.23\textwidth}
         \centering
    \includegraphics[width=\textwidth]{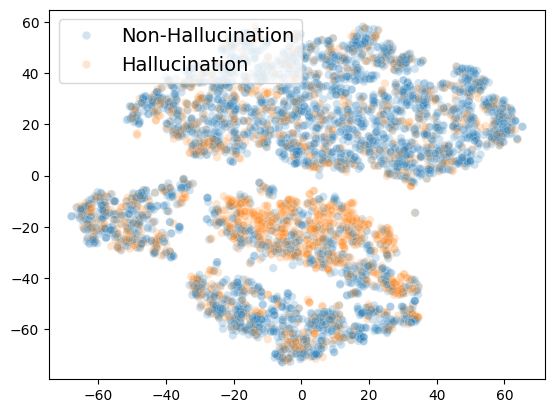}
         \caption{Self-attention scores}
     \end{subfigure}
    \hfill
     \begin{subfigure}[b]{0.23\textwidth}
         \centering
    \includegraphics[width=\textwidth]{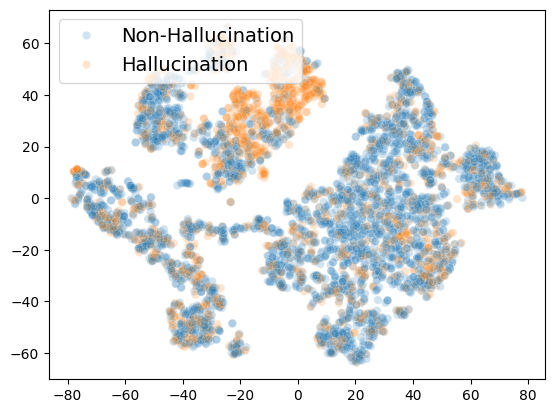}
         \caption{Fully-connected activations}
     \end{subfigure}
    \caption{[\falconM on TriviaQA dataset] 2D TSNE projections of the model artifacts for hallucinations and non-hallucinations.
    }
    \label{fig:entropy_main_plot}
\end{figure*}

\subsection{Models}
We analyze responses from three different models: OpenLLaMA ({\tt LAM}),%
\footnote{\url{https://huggingface.co/docs/transformers/main/model_doc/open-llama}}
OPT ({\tt OPT})%
\footnote{\url{https://huggingface.co/docs/transformers/v4.19.2/en/model_doc/opt}}
and Falcon ({\tt FAL}).%
\footnote{\url{https://huggingface.co/docs/transformers/main/model_doc/falcon}}
All three models come with different size-based variants. We consider two different sizes for each model: \llamaS (7 billion parameters) and \llamaM; \optS and \optM; \falconS and \falconM. We consider different variants to study the effect of model size on the hallucination detection performance.

\subsection{Infrastructure and Parameters} \label{sec:params}
All experiments were ran on an Amazon SageMaker {\tt ml.g5.12xlarge} instance with $192$ GB RAM and 4x NVIDIA A10G GPUs. Self-attention and fully-connected activations were captured using Amazon SageMaker Debugger \cite{rauschmayr2021debugger}.

Generations are performed by sampling the most likely token according to the Softmax probability. This corresponds to using a Temperature = 0. We continue generating until an \texttt{<end of text>} tokens is generated or the generation is $100$ tokens long.

We compute the Integrated Gradients attribution using Captum~\cite{kokhlikyan2020captum}. We use a baseline of all $0$s and the number of IG iterations is set of $50$. Given an output token, the corresponding IG attributions for each input token are a vector of the same dimensionality as the input token embeddings. We convert the vector scores to the a per token scalar score by using the L2 norm reduction, which has been shown to provide similar or better performance to other reduction strategies~\cite{zafar2021lack}.

Hallucination classifiers are trained with a batch size of $128$ for $1,000$ iterations. We use Adam optimizer with a learning rate of $10^{-4}$ and weight decay of $10^{-2}$.

\subsection{Question Answering Accuracy}\label{sec:qa_accuracy}

Before moving on to hallucination detection, we first analyze the performance of the models in correctly answering the questions \ie how often the models hallucinate.
Table~\ref{table:qa_accuracy} shows the accuracy of different models.
The models showed a range of performance across each task. All models performed best on the TriviaQA tasks, and worst on the Birth Place task.
Surprisingly, larger models did not always perform better. On the \textit{subject specific tasks} from \trex, smaller models often performed as well or better than their larger counterparts (6 out of 9 times). On the more \textit{general TriviaQA task}, larger models consistently performed better. On all tasks, \falconM significantly outperformed all other models.
Further, while larger models on average performed better at general knowledge tasks, variation in performance is more strongly correlated with model type rather than size. For example, while \llamaM outperformed its smaller variants \llamaS, and similarly \optM outperformed its smaller variant, both {\tt LAM} models outperformed both {\tt OPT} models.

\subsection{Accuracy of the hallucination heuristic}
\label{sec:heuristic_eval}

Recall from \S\ref{sec:setup} that we consider a model answer $\answer$ to be a hallucination if $\refanswer \not \subseteq \answer$. To evaluate this heuristic, we randomly sample $100$  \triviaqa generations with \falconM. For each generation, each of the three authors independently labeled it as (non-) hallucination. Based on the majority vote, \textbf{the heuristic was correct in $98 / 100$ cases}. The cases where the heuristic was incorrect are:

\begin{tcolorbox}[
    fontupper=\ttfamily,
    left=1pt,
    right=1pt,
    top=1pt,
    bottom=1pt,
    standard jigsaw,
    opacityback=0
]
\textbf{Question:} In Greek mythology, one of the 12 Labours of Hercules was to produce ‘what’ item belonging to Amazonian queen Hippolyte?
\\
\textbf{Ref. Answer:} Magical Girdle
\\
\textbf{Model Answer:} The girdle of Hippolyte was a girdle that was worn by the Amazonian queen Hippolyte. It was one of the 12 Labours of Hercules.
\\
\textbf{Heuristic label:} Hallucination
\\
\textbf{Human label:} Not a hallucination
\\
\\
\textbf{Question:} Which British Prime Minister was the 1st Earl of Stockton?
\\
\textbf{Ref. Answer:} Harold Macmillan or Earl of Stockton
\\
\textbf{Model Answer:} Which British Prime Minister was the 1st Earl of Stockton?
\\
\textbf{Heuristic label:} Not a hallucination
\\
\textbf{Human label:} Hallucination
\end{tcolorbox}

\begin{table*}[ht]
\centering
\begin{subtable}{.45\linewidth}
\centering
\begin{tabular}{l  c  c  c  c  c  c} 
\toprule
 {} & \textbf{\llamaM} & \textbf{\optM} & \textbf{\falconM} \\ 
 \midrule
 \textbf{TriviaQA} & 0.60 & 0.57 & 0.48 \\ 
 \textbf{Capitals} & 0.41 & 0.68 & 0.43 \\
 \textbf{Founders} & 0.57 & 0.44 & 0.45 \\
 \textbf{Birth Place} & 0.48 & 0.45 & 0.51 \\
 \textbf{Combined} & 0.43 & 0.57 & 0.49 \\
 \bottomrule
\end{tabular}
 \caption{IG attributions}
\end{subtable}
\hspace{0.07\linewidth}
\begin{subtable}{.45\linewidth}
\centering
\begin{tabular}{l  c  c  c  c  c  c} 
 \toprule
{} & \textbf{\llamaM} & \textbf{\optM} &
\textbf{\falconM} \\ 
\midrule
 \textbf{TriviaQA} & 0.71 & 0.63 & 0.60 \\ 
 \textbf{Capitals} & 0.68 & 0.69 & 0.63 \\
 \textbf{Founders} & 0.61 & 0.67 & 0.66 \\
 \textbf{Birth Place} & 0.66 & 0.62 & 0.66 \\
 \textbf{Combined} & 0.69 & 0.67 & 0.62 \\
 \bottomrule
\end{tabular}
\caption{Softmax probabilities}
\end{subtable}
\\
\vspace{0.5cm}
\begin{subtable}{.45\linewidth}
\centering
\begin{tabular}{ l  c  c  c  c c  c} 
 \toprule
{} & \textbf{\llamaM} & \textbf{\optM} &
\textbf{\falconM} \\ 
\midrule
 \textbf{TriviaQA} & 0.71 & 0.65 & 0.71 \\ 
 \textbf{Capitals} & 0.72 & 0.72 & 0.72 \\
 \textbf{Founders} & 0.73 & 0.68 & 0.71 \\
 \textbf{Birth Place} & 0.81 & 0.61 & 0.81\\
 \textbf{Combined} & 0.78 & 0.71 & 0.79\\
 \bottomrule
\end{tabular}
\caption{Self-attention scores}
\end{subtable}
\hspace{0.07\linewidth}
\begin{subtable}{.45\linewidth}
\centering
\begin{tabular}{ l  c  c  c  c c  c} 
 \toprule
{} & \textbf{\llamaM} & \textbf{\optM} &
\textbf{\falconM} \\ 
\midrule
 \textbf{TriviaQA} & 0.72 & 0.64 & 0.72 \\ 
 \textbf{Capitals} & 0.73 & 0.71 & 0.70 \\
 \textbf{Founders} & 0.71 & 0.72 & 0.73 \\
 \textbf{Birth Place} & 0.80 & 0.77 & 0.76\\
 \textbf{Combined} & 0.79 & 0.73 & 0.82\\
 \bottomrule
\end{tabular}
\caption{Fully-connected activations}
\end{subtable}
\caption{[Larger model variants] Test AUROC of binary classifiers in detecting hallucinations. }
\label{table:rocauc_large_models}
\end{table*}

\begin{table*}[ht]
\centering
\begin{subtable}{.45\linewidth}
\centering
\begin{tabular}{l  c  c  c  c  c  c} 
\toprule
 {} & \textbf{\llamaS} & \textbf{\optS} & \textbf{\falconS} \\ 
 \midrule
 \textbf{TriviaQA} & 0.62 & 0.54 & 0.54 \\ 
 
 \textbf{Capitals} & 0.35 & 0.34 & 0.40 \\
 
 \textbf{Founders} & 0.40 & 0.48 & 0.54 \\
 
 \textbf{Birth Place} & 0.42 & 0.44 & 0.43 \\
 
 \textbf{Combined} & 0.40 & 0.42 & 0.53 \\
 
 \bottomrule
\end{tabular}
\caption{IG attributions}
\end{subtable}
\hspace{0.07\linewidth}
\begin{subtable}{.45\linewidth}
\centering
\begin{tabular}{l  c  c  c  c  c  c} 
 \toprule
{} & \textbf{\llamaS} & \textbf{\optS} & \textbf{\falconS} \\ 
\midrule
 \textbf{TriviaQA} & 0.68 & 0.65 & 0.64 \\
 
 \textbf{Capitals} & 0.65 & 0.67 & 0.68 \\
 \textbf{Founders} & 0.66 & 0.66 & 0.72 \\
 \textbf{Birth Place} & 0.59 & 0.65 & 0.64 \\
 \textbf{Combined} & 0.67 & 0.67 & 0.72 \\
 \bottomrule
\end{tabular}
\caption{Softmax probabilities}
\end{subtable}
\\
\vspace{0.5cm}
\begin{subtable}{.45\linewidth}
\centering
\begin{tabular}{l  c  c  c  c  c  c} 
 \toprule
{} & \textbf{\llamaS} & \textbf{\optS} & \textbf{\falconS} \\ 
\midrule
 \textbf{TriviaQA} & 0.71 & 0.66 & 0.66 \\ 
 
 \textbf{Capitals} & 0.76 & 0.73 & 0.68 \\
 
 \textbf{Founders} & 0.70 & 0.72 & 0.66 \\
 
 \textbf{Birth Place} & 0.76 & 0.64 & 0.72 \\
 
 \textbf{Combined} & 0.75 & 0.75 & 0.71 \\
 
 \bottomrule
\end{tabular}
\caption{Self-attention scores}
\end{subtable}
\hspace{0.07\linewidth}
\begin{subtable}{.45\linewidth}
\centering
\begin{tabular}{ l  c  c  c  c c  c} 
 \toprule
{} & \textbf{\llamaS} & \textbf{\optS} & \textbf{\falconS} \\ 
\midrule
 \textbf{TriviaQA} & 0.74 & 0.70 & 0.70 \\ 
 
 \textbf{Captials} & 0.74 & 0.75 & 0.73 \\
 
 \textbf{Founders} & 0.72 & 0.71 & 0.70 \\
 
 \textbf{Birth Place} & 0.71 & 0.71 & 0.78 \\
 
 \textbf{Combined} & 0.77 & 0.77 & 0.71 \\
 
 \bottomrule
\end{tabular}
\caption{Fully-connected activations}
\end{subtable}
\caption{[Smaller model variants] Test AUROC of binary classifiers in detecting hallucinations. }
\label{table:rocauc_small_models}
\end{table*}

\section{Results}

We first qualitatively investigate the differences in distributions of hallucinations and non-hallucinations and then quantitatively evaluate the performance of hallucination classifiers.

\subsection{Qualitative analysis} \label{sec:qualitative}

Recall our hypotheses that hallucinating and non-hallucinating generations differ have different distributions of generation artifacts, namely, Softmax probabilities, IG attributions, self attention and fully-connected activations (\S\ref{sec:artifacts}).
Examining \falconM on the TriviaQA dataset (Figure~\ref{fig:entropy_main_plot}), we find that hallucinated outputs indeed tend to be distributed differently from non-hallucinated outputs when considering Softmax probabilities, self-attention scores and fully-connected activations. The IG attributions however do not show much difference in distributions.

\begin{figure}[ht!]
    \centering
    \includegraphics[width=\columnwidth]{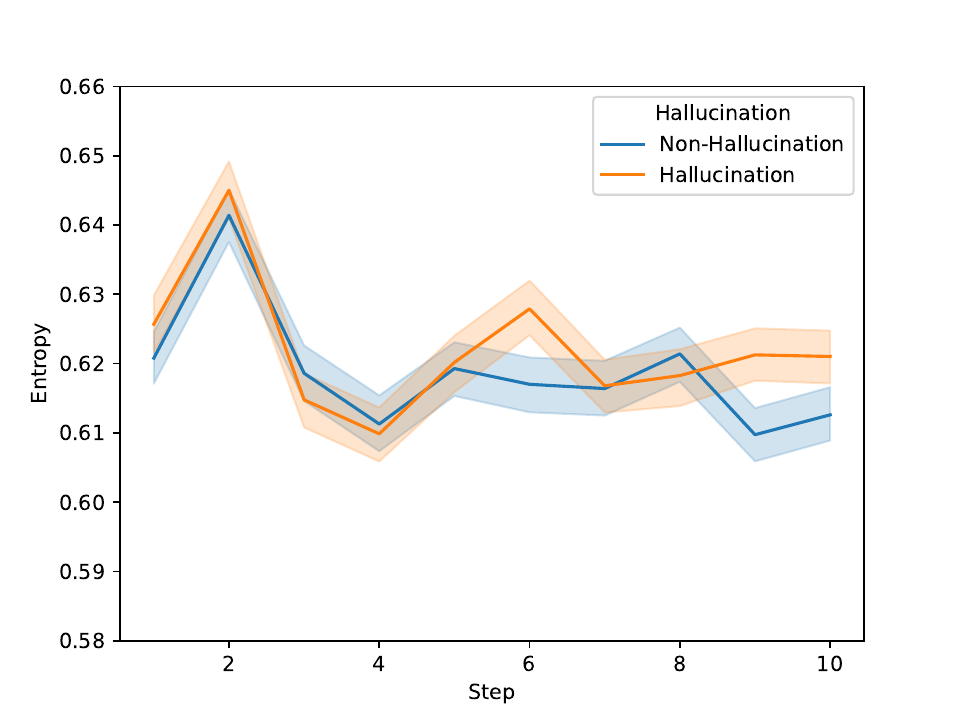}
    \caption{[\falconM on \triviaqa dataset] Softmax entropy at different generation steps. The difference in entropy between hallucinated and non-hallucinated outputs does not vary much with a change in the generation steps.}
    \label{fig:falcon_softmax_entropy_all_tokens}
\end{figure}

\begin{figure}[ht]
    \centering
    \includegraphics[width=\columnwidth]{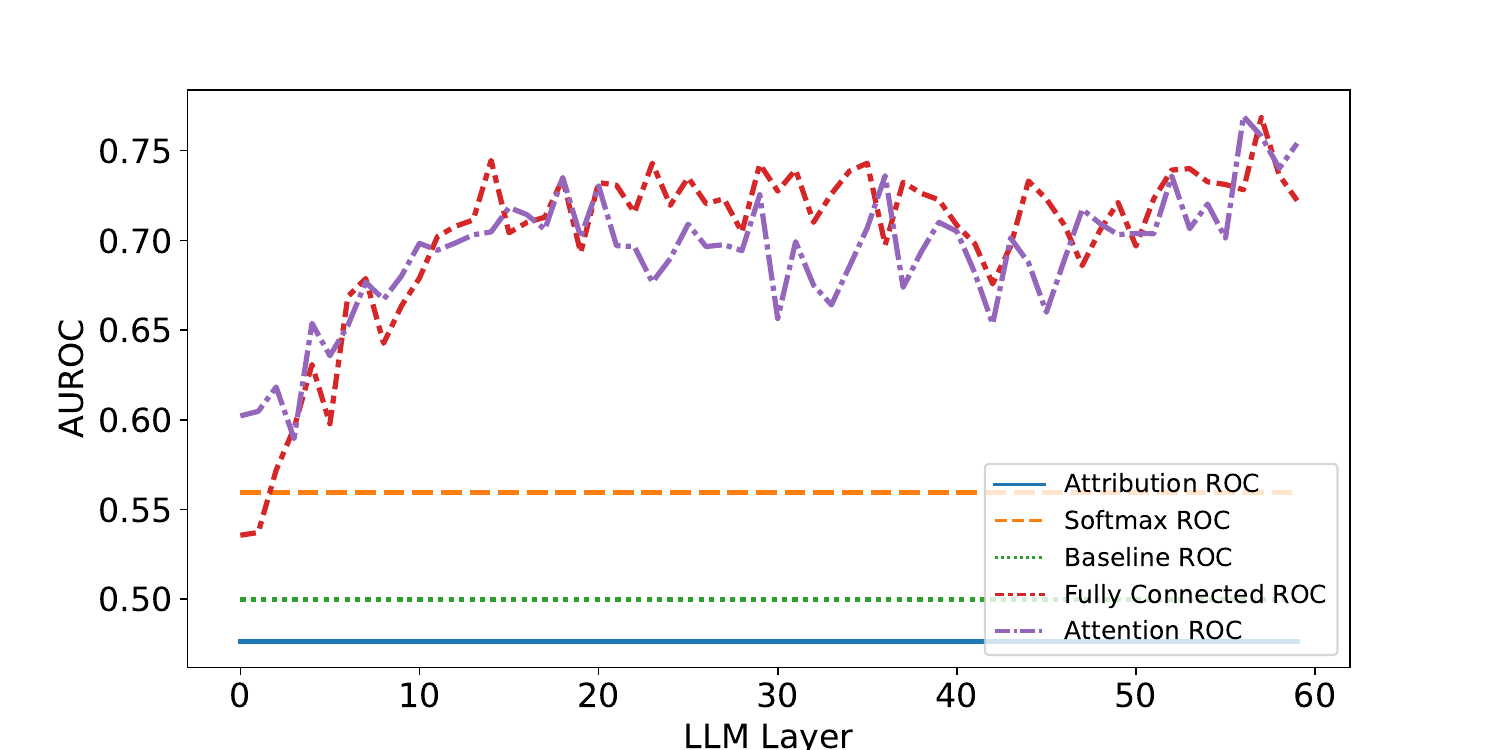}
    \caption{[\falconM on \triviaqa dataset] AUROC of the hallucination detectors using self-attention and fully-connected activations at different layers. The performance is better at later layers but has diminishing returns. }
    \label{fig:layerwise_roc}
\end{figure}

The results when comparing the entropy of distributions, and other datasets and models are mixed (figures omitted due to lack of space). The distributions of input token entropy showed a stronger correlation to dataset type than to model. For topic specific datasets, entropy is on average lower for hallucinated results, while it is slightly higher for hallucinated results on TriviaQA. The entropies of the Softmax outputs are less stable across both datasets and models. While most cases show slightly higher entropy for hallucinated results, the difference is inconsistent. Further, in both the input token and output Softmax cases, the differences, while present, are relatively small. We find similarly small difference in entropy along all generated tokens (see Figure~\ref{fig:falcon_softmax_entropy_all_tokens}). 2-dimensional TSNE plots also show similarly mixed trends.

\textbf{While mixed, the results show that in many cases, the reduced distributions of model artifacts (either to 2 dimensions via TSNE or to 1 dimension via entropy) show visually discernible differences between hallucinated and non-hallucinated outputs}. Next we investigate if we can train accurate hallucination detectors by using these artifacts in their original form \ie without reductions.

\subsection{Hallucination Classifiers}\label{sec:quantitative}

Tables~\ref{table:rocauc_large_models} and~\ref{table:rocauc_small_models} shows the test AUROC when detecting hallucinations. We opt for AUROC instead of binary classification accuracy because of high class imbalance (Table~\ref{table:qa_accuracy}). We include the results for binary classification accuracy in Appendix~\ref{app:additional_results}.

Results show that the IG attribution does slightly better than random chance on the TriviaQA dataset, but no better than random on the subject specific datasets. Softmax consistently does better than random for all tasks.
Self-attention scores and fully-connected activations outperform both IG and Softmax. Interestingly, these results hold even though $96\%$ of generated responses start with the newline character, indicating that the classifier is not simply learning tokens that correlate with hallucination, but that even with the same token, the model internal state between hallucinated and non-hallucinated results differs.

While overall accuracy correlates with model size within each model type, the trend is less consistent for a given model's ability to identify its own hallucinations. In the case of {\tt LAM}, the larger variant consistently performed better at identifying hallucination through our classifier. {\tt OPT} and {\tt FAL}, on the other hand, showed no consistent correlation to size in their ability to capture hallucinations. In both the large and small model variants, the fully-connected and self attention activation internal states provided the best performance at identifying hallucinations, followed by the model's Softmax output.

A classifier trained on all four datasets combined (labeled as ``Combined'' in Tables~\ref{table:rocauc_large_models} and \ref{table:rocauc_small_models})  tended to perform slightly better than the individual datasets. However, a classifier trained on all model artifacts  combined  (Softmax, IG, self-attention and fully-connected activations) showed no improvement beyond models trained on each artifact individually (Table~\ref{table:combined_artifacts}). There could be several reasons for this: the input dimensionality of this combined classifier  is much higher than the individual classifiers and it is architecturally more complex as it consists of both dense and recurrent units. Training this mixed architecture might require special considerations. We leave the detailed analysis to a follow up work.

In summary we note that different model artifacts provide different level of accuracy in detecting hallucinations and \textbf{in most cases, self-attention scores and fully-connected activations provide over $0.70$ AUROC in detecting hallucinations over a range of datasets and models}. 

\subsection{Hyperparameters and baseline comparison} \label{sec:ablations}

We take a closer look at various hyperparameters and also conduct a performance comparison with an existing method.

\xhdr{Effect of Transformer layer choice}
The results in \S\ref{sec:quantitative} were based on using the self-attention and fully-connected activations from the last Transformer layer. We also investigate how the performance would change as a results of a change in layer number.
Results in Figure~\ref{fig:layerwise_roc} show that the performance of our classifier improves with depth of the model layers. Early layers do only slightly better than random chance, while later layers shows significant improvement beyond random chance. 

\xhdr{Effect of classifier hyperparameters}
Given the simplicity of our hallucination detection classifiers, the only hyperparameters available were batch size, learning rate, and weight decay. We hand-selected the defaults in \S\ref{sec:params}. We also tried (\triviaqa with \falconM) a range of different options for these hyperparameters where we varied batch size from $4$ to $256$, learning rate from $10^{-6}$ to $10^{-2}$ and weight decay from $10^{-4}$ to $10^{-1}$. We found that the smallest batch size and largest learning rate performed worse. But smaller adjustments (batch size $128$ vs $256$, learning rate $1 \times 10^{-4}$ \vs $2 \times 10^{-4}$) made no discernible difference in performance.

We also tested larger models for both our GRU and MLP architectures. On the GRU, we tested from $4$ to $12$ gated recurrent layers. On our MLP, we tested up to $8$ layers, with widths from $32$ to $256$. In both cases, we found that larger models provided no benefit in either AUROC or accuracy. For example, for the GRU model \falconM with \triviaqa, expanding to $12$ recurrent layers yielded an AUC of $0.46$, slightly worse than the $0.48$ reported with $4$ layers reported in Table~\ref{table:rocauc_large_models}. For MLP, expanding to $8$ layers yielded $0.72$, only slightly better than the $0.71$ we reported with the single layer model.

A systematic study of these hyperparameter choices, and understanding the effect of more advanced detector architectures (\eg Transformers, LSTMs) on the detection performance is a promising future direction.

\begin{table}[t]
\centering
\begin{tabular}{ l  c  c  c  c c  c} 
 \toprule
{} & \textbf{\llamaM} & \textbf{\optM} &
\textbf{\falconM} \\ 
\midrule
 \textbf{TriviaQA} & 0.75 & 0.64 & 0.65 \\ 
 \textbf{Capitals} & 0.69 & 0.55 & 0.62 \\
 \textbf{Founders} & 0.66 & 0.53 & 0.63 \\
 \textbf{Birth Place} & 0.53 & 0.59 & 0.68\\
 \textbf{Combined} & 0.58 & 0.58 & 0.60\\
 \bottomrule
\end{tabular}
\caption{Test AUROC of the classifier trained on all four artifacts at once. The performance is not better than the case when individual artifacts are considered (Tables~\ref{table:rocauc_large_models} and~\ref{table:rocauc_small_models}).}
\label{table:combined_artifacts}
\end{table}

\xhdr{Comparison with SelfCheckGPT}
We compare the performance of our method to SelfCheckGPT~\cite{manakul2023selfcheckgpt}. Results are shown in Table~\ref{table:selfcheckgpt_large} for larger model variants and Table~\ref{table:selfcheckgpt_small} for smaller model variants. The performance is generally worse than our classifiers. We use two variants of SelfCheckGPT: BERTScore and n-gram. We use $20$ generations per input and set the temperature to $0.1$. Preliminary analysis in Table~\ref{table:selfcheckgpt_temp1} shows that setting the temperature to $1.0$ as suggested in the paper~\cite{manakul2023selfcheckgpt} does not lead to better performance.

\begin{table*}[t]
\centering
\begin{subtable}{.45\linewidth}
\centering
\begin{tabular}{l  c  c  c  c  c  c} 
\toprule
 {} & \textbf{\llamaM} & \textbf{\optM} & \textbf{\falconM} \\ 
 \midrule
 \textbf{TriviaQA} & 0.56 & 0.58 & 0.50 \\ 
 \textbf{Capitals} & 0.53 & 0.51 & 0.56 \\
 \textbf{Founders} & 0.55 & 0.49 & 0.57 \\
 \textbf{Birth Place} & 0.51 & 0.54 & 0.48 \\
 \bottomrule
\end{tabular}
 \caption{BERTScore}
\end{subtable}
\hspace{0.07\linewidth}
\begin{subtable}{.45\linewidth}
\centering
\begin{tabular}{l  c  c  c  c  c  c} 
 \toprule
{} & \textbf{\llamaM} & \textbf{\optM} &
\textbf{\falconM} \\ 
\midrule
 \textbf{TriviaQA} & 0.54 & 0.53 & 0.54 \\ 
 \textbf{Capitals} & 0.54 & 0.55 & 0.54 \\
 \textbf{Founders} & 0.58 & 0.56 & 0.57 \\
 \textbf{Birth Place} & 0.51 & 0.49 & 0.52 \\
 \bottomrule
\end{tabular}
\caption{n-gram}
\end{subtable}
\caption{[Larger model variants] AUROC of SelfCheckGPT based on BERTScore and n-gram scores.}
\label{table:selfcheckgpt_large}
\end{table*}

\begin{table*}[t]
\centering
\begin{subtable}{.45\linewidth}
\centering
\begin{tabular}{l  c  c  c  c  c  c} 
\toprule
 {} & \textbf{\llamaS} & \textbf{\optS} & \textbf{\falconS} \\ 
 \midrule
 \textbf{TriviaQA} & 0.55 & 0.56 & 0.55 \\ 
 \textbf{Capitals} & 0.52 & 0.50 & 0.49 \\
 \textbf{Founders} & 0.58 & 0.58 & 0.54 \\
 \textbf{Birth Place} & 0.52 & 0.51 & 0.56 \\
 \bottomrule
\end{tabular}
 \caption{BERTScore}
\end{subtable}
\hspace{0.07\linewidth}
\begin{subtable}{.45\linewidth}
\centering
\begin{tabular}{l  c  c  c  c  c  c} 
 \toprule
{} & \textbf{\llamaS} & \textbf{\optS} &
\textbf{\falconS} \\ 
\midrule
 \textbf{TriviaQA} & 0.53 & 0.58 & 0.50 \\ 
 \textbf{Capitals} & 0.53 & 0.48 & 0.49 \\
 \textbf{Founders} & 0.53 & 0.43 & 0.46 \\
 \textbf{Birth Place} & 0.57 & 0.58 & 0.55 \\
 \bottomrule
\end{tabular}
\caption{n-gram}
\end{subtable}
\caption{[Smaller model variants] AUROC of SelfCheckGPT based on BERTScore and n-gram scores.}
\label{table:selfcheckgpt_small}
\end{table*}

\begin{table}[ht]
\centering
\begin{tabular}{l  c  c  c  c  c  c} 
\toprule
 {} & \textbf{BERTScore} & \textbf{n-gram} \\ 
 \midrule
 \textbf{TriviaQA} & 0.52 & 0.55 \\ 
 \textbf{Capitals} & 0.47 & 0.50 \\
 \textbf{Founders} & 0.44 & 0.51 \\
 \textbf{Birth Place} & 0.55 & 0.48 \\
 \bottomrule
\end{tabular}
\vspace{2mm}
 \caption{[\falconS] AUROC of SelfCheckGPT with $T = 1$.}
 \label{table:selfcheckgpt_temp1}
\end{table}

\section{Conclusion, Discussion \& Limitations} \label{sec:conclusion}

We address the problematic behavior where large language models hallucinate incorrect facts.
By extracting different generation artifacts such as Integrated Gradients feature attribution scores and self-attention scores at various Transformer layers we build simple classifiers for detecting hallucinations. The detection can already be performed at the start of the response generation, that is, factual hallucinations can often be detected even before they occur. We show that the ability to identify hallucinations persists across different topics like Capitals, Founders and Birth Place; and in a broader trivia knowledge context. Attaching classifiers of this type to deployed LLMs can serve as an effective method of flagging potentially incorrect information before it reaches the end-users or downstream applications.

Surprisingly, the classifiers are able to detect hallucinations even when the first generated token is a seemingly ``uninformative'' token like a formatting character. With the exception of Falcon models on the Birth Place dataset, the newline character is the first token in $99.5$\% of generated responses, with the remaining $0.5$\% being a punctuation. In the Falcon / Birth Place case, the newline character starts $82$\% of responses, while the word ``where'' starts the other $18$\%.
Two factors likely contribute to this surprising effectiveness of such ``uninformative'' tokens. First, while the first token {\tt \textbackslash n} is a single character, the associated artifacts like hidden states and softmax probabilities contain much more information as these are vectors with thousands of dimensions. Second, we are experimenting with autoregressive models. At the first generation location, the model has ingested all the external information (that is, the input) and the upcoming generation only depends on the internals of the model itself (minus the effect of the generation strategy which in our case is the most likely next token). Nonetheless, this finding merits further investigation and is a promising avenue for future work. 

We only considered factual hallucinations in this work. It would be interesting to study if the proposed method also extends to other forms of hallucinations~\cite{huang2023survey,ji2023survey}. We also considered a rather simple setup where the model output is supposed to contains a single fact only. Extending our method to probe multiple hallucinations (\eg multiple facts in a biography like place of birth, date of birth, alma mater) in a single generation is also an important follow up direction.

Our method cannot be used with ``blackbox'' models that do not reveal model internals and gradients to the users. Text-only methods like SelfCheckGPT are well-suited for such situations but compensate for the lack of internal access by querying the model several times. In our case, all except one artifact can be computed ``for free'' during the forward pass.  Nonetheless, our method can still be deployed by the model providers internally. 
Experimenting with more elaborate retrieval settings (\eg those using retrieval augmented generation~\cite{lewis2020retrieval}, instruction tuned models~\cite{ouyang2022training} and in-context examples~\cite{wu2024towards}), a wider range of datasets and model types is also a promising avenue for future work. 

Finally, the goal of this paper was not to extract the maximum possible detection performance, but to test if certain artifacts can be promising in detecting hallucinations. For this reason, we use relatively simple detection architectures. We leave a more in-depth analysis of detection architectures to a future study.

The code repository for the paper is available at: \url{https://github.com/amazon-science/llm-hallucinations-factual-qa}.

\printbibliography

\appendix

\section{Additional Results} \label{app:additional_results}
Recall that due to the imbalance between hallucinations vs. non-hallucinations (Section~\ref{sec:qa_accuracy}), we reported AUROC in the main experiments (Section~\ref{sec:quantitative}). Tables~\ref{table:binary_acc_large_models} and~\ref{table:binary_acc_small_models} show the binary classification accuracy for larger and smaller model variants.

\begin{table*}[ht]
\centering
\begin{subtable}{.45\linewidth}
\centering
\begin{tabular}{l  c  c  c  c  c  c} 
\toprule
 {} & \textbf{\llamaM} & \textbf{\optM} & \textbf{\falconM} \\ 
 \midrule
 \textbf{TriviaQA} & 0.64 & 0.67  & 0.66  \\ 
 \textbf{Capitals} & 0.68 & 0.65  & 0.63  \\
 \textbf{Founders} & 0.70 & 0.69  & 0.68  \\
 \textbf{Birth Place} & 0.85 & 0.87  & 0.89 \\
 \bottomrule
\end{tabular}
 \caption{IG attributions}
\end{subtable}
\hspace{0.07\linewidth}
\begin{subtable}{.45\linewidth}
\centering
\begin{tabular}{l  c  c  c  c  c  c} 
 \toprule
{} & \textbf{\llamaM} & \textbf{\optM} &
\textbf{\falconM} \\ 
\midrule
 \textbf{TriviaQA} & 0.65  & 0.63  & 0.68  \\ 
 \textbf{Capitals} & 0.70  & 0.71  & 0.63  \\
 \textbf{Founders} & 0.73  & 0.68  & 0.67  \\
 \textbf{Birth Place} & 0.86  & 0.91  & 0.92  \\
 \bottomrule
\end{tabular}
\caption{Softmax probabilities}
\end{subtable}
\\
\vspace{0.5cm}
\begin{subtable}{.45\linewidth}
\centering
\begin{tabular}{ l  c  c  c  c c  c} 
 \toprule
{} & \textbf{\llamaM} & \textbf{\optM} &
\textbf{\falconM} \\ 
\midrule
 \textbf{TriviaQA} & 0.69  & 0.65  & 0.70  \\ 
 \textbf{Capitals} & 0.70  & 0.71  & 0.66  \\
 \textbf{Founders} & 0.72  & 0.74  & 0.68  \\
 \textbf{Birth Place} & 0.89  & 0.91  & 0.93 \\
 \bottomrule
\end{tabular}
\caption{Self-attention scores}
\end{subtable}
\hspace{0.07\linewidth}
\begin{subtable}{.45\linewidth}
\centering
\begin{tabular}{ l  c  c  c  c c  c} 
 \toprule
{} & \textbf{\llamaM} & \textbf{\optM} &
\textbf{\falconM} \\ 
\midrule

 \textbf{TriviaQA} & 0.67  & 0.65  & 0.69  \\ 
 \textbf{Capitals} & 0.69  & 0.70  & 0.65  \\
 \textbf{Founders} & 0.73  & 0.74  & 0.68  \\
 \textbf{Birth Place} & 0.88  & 0.89  & 0.94 \\
 
 \bottomrule
\end{tabular}
\caption{Fully-connected activations}
\end{subtable}
\caption{[Larger model variants] Test accuracy of binary classifiers in detecting hallucinations. }
\label{table:binary_acc_large_models}
\end{table*}

\begin{table*}[ht]
\centering
\begin{subtable}{.45\linewidth}
\centering
\begin{tabular}{l  c  c  c  c  c  c} 
\toprule
 {} & \textbf{\llamaS} & \textbf{\optS} & \textbf{\falconS} \\ 
 \midrule
 \textbf{TriviaQA} & 0.63  & 0.68  & 0.63  \\ 
 \textbf{Capitals} & 0.70  & 0.64  & 0.67  \\
 \textbf{Founders} & 0.69  & 0.70  & 0.68  \\
 \textbf{Birth Place} & 0.83  & 0.88  & 0.89 \\
 \bottomrule
\end{tabular}
 \caption{IG attributions}
\end{subtable}
\hspace{0.07\linewidth}
\begin{subtable}{.45\linewidth}
\centering
\begin{tabular}{l  c  c  c  c  c  c} 
 \toprule
{} & \textbf{\llamaS} & \textbf{\optS} &
\textbf{\falconS} \\ 
\midrule
 \textbf{TriviaQA} & 0.66  & 0.63  & 0.66  \\ 
 \textbf{Capitals} & 0.67  & 0.68  & 0.64  \\
 \textbf{Founders} & 0.71  & 0.63  & 0.68  \\
 \textbf{Birth Place} & 0.87  & 0.90  & 0.89  \\
 \bottomrule
\end{tabular}
\caption{Softmax probabilities}
\end{subtable}
\\
\vspace{0.5cm}
\begin{subtable}{.45\linewidth}
\centering
\begin{tabular}{ l  c  c  c  c c  c} 
 \toprule
{} & \textbf{\llamaS} & \textbf{\optS} &
\textbf{\falconS} \\ 
\midrule
 \textbf{TriviaQA} & 0.65  & 0.62  & 0.72  \\ 
 \textbf{Capitals} & 0.70  & 0.72  & 0.65  \\
 \textbf{Founders} & 0.69  & 0.71  & 0.69  \\
 \textbf{Birth Place} & 0.91  & 0.90  & 0.91 \\
 \bottomrule
\end{tabular}
\caption{Self-attention scores}
\end{subtable}
\hspace{0.07\linewidth}
\begin{subtable}{.45\linewidth}
\centering
\begin{tabular}{ l  c  c  c  c c  c} 
 \toprule
{} & \textbf{\llamaS} & \textbf{\optS} &
\textbf{\falconS} \\ 
\midrule

 \textbf{TriviaQA} & 0.66  & 0.68  & 0.66  \\ 
 \textbf{Capitals} & 0.70  & 0.69  & 0.65  \\
 \textbf{Founders} & 0.71  & 0.72  & 0.70  \\
 \textbf{Birth Place} & 0.87  & 0.88  & 0.93  \\
 
 \bottomrule
\end{tabular}
\caption{Fully-connected activations}
\end{subtable}
\caption{[Smaller model variants] Test accuracy of binary classifiers in detecting hallucinations. }
\label{table:binary_acc_small_models}
\end{table*}

\begin{figure*}[ht]
    \centering
    \includegraphics[width=\textwidth]{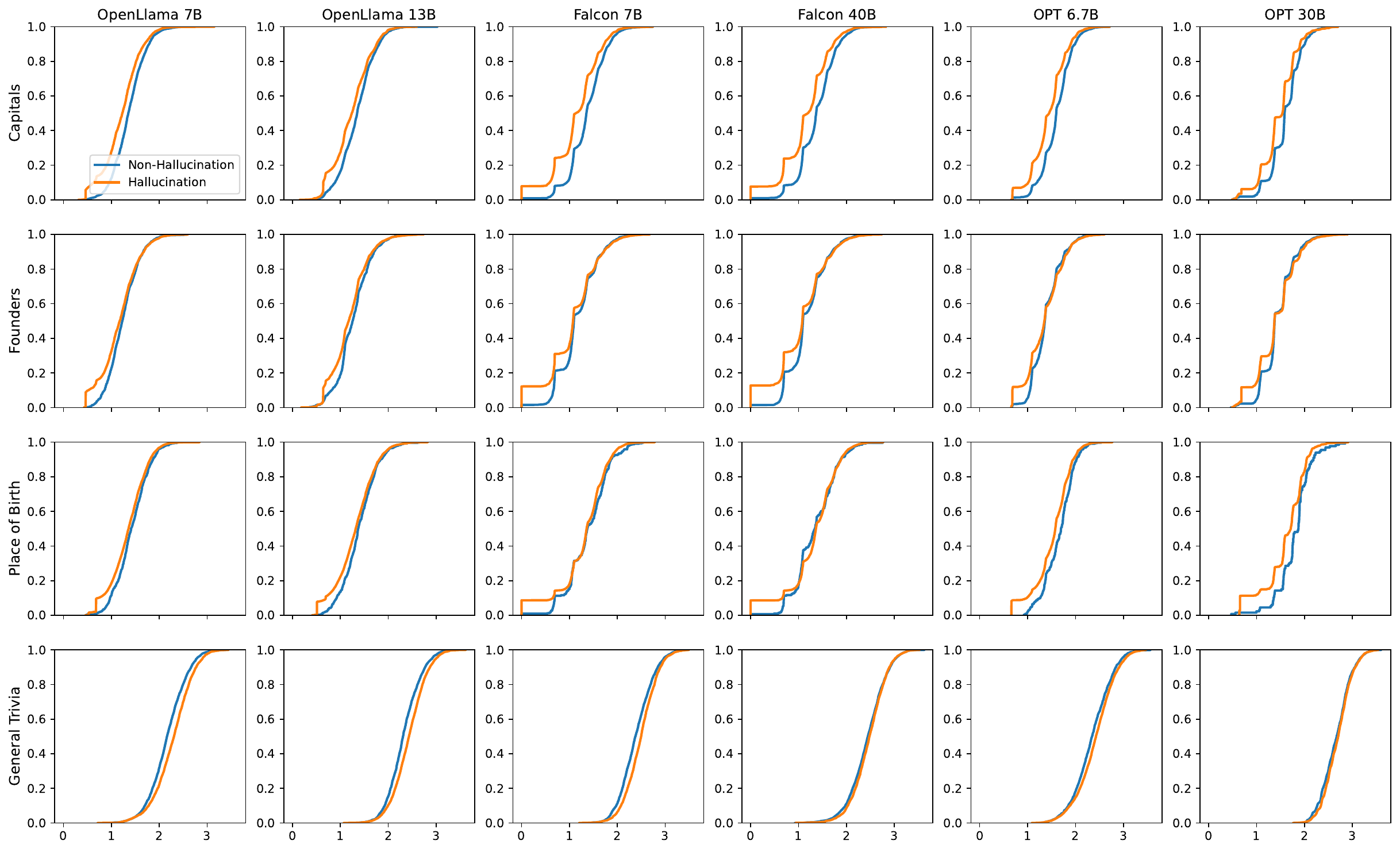}
    \caption{Cumulative distribution function of entropy in IG attributions for all models and datasets.}
    \label{fig:ig_entropy_all}
\end{figure*}
\begin{figure*}[ht]
    \centering
    \includegraphics[width=\textwidth]{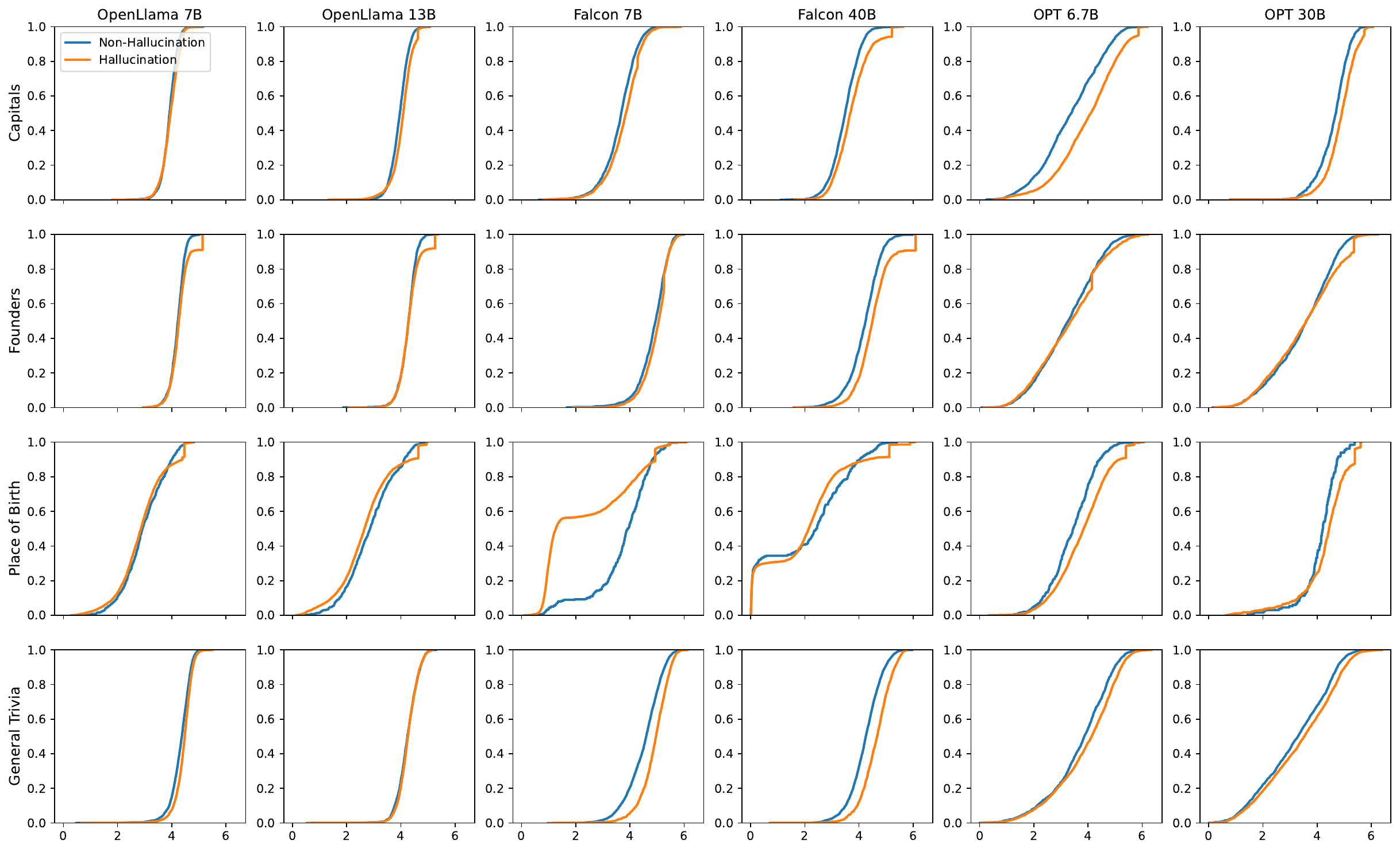}
    \caption{Cumulative distribution function of entropy in Softmax probabilities for all models and datasets.}
    \label{fig:softmax_entropy_all}
\end{figure*}
\begin{figure*}[ht]
    \centering
    \includegraphics[width=\textwidth]{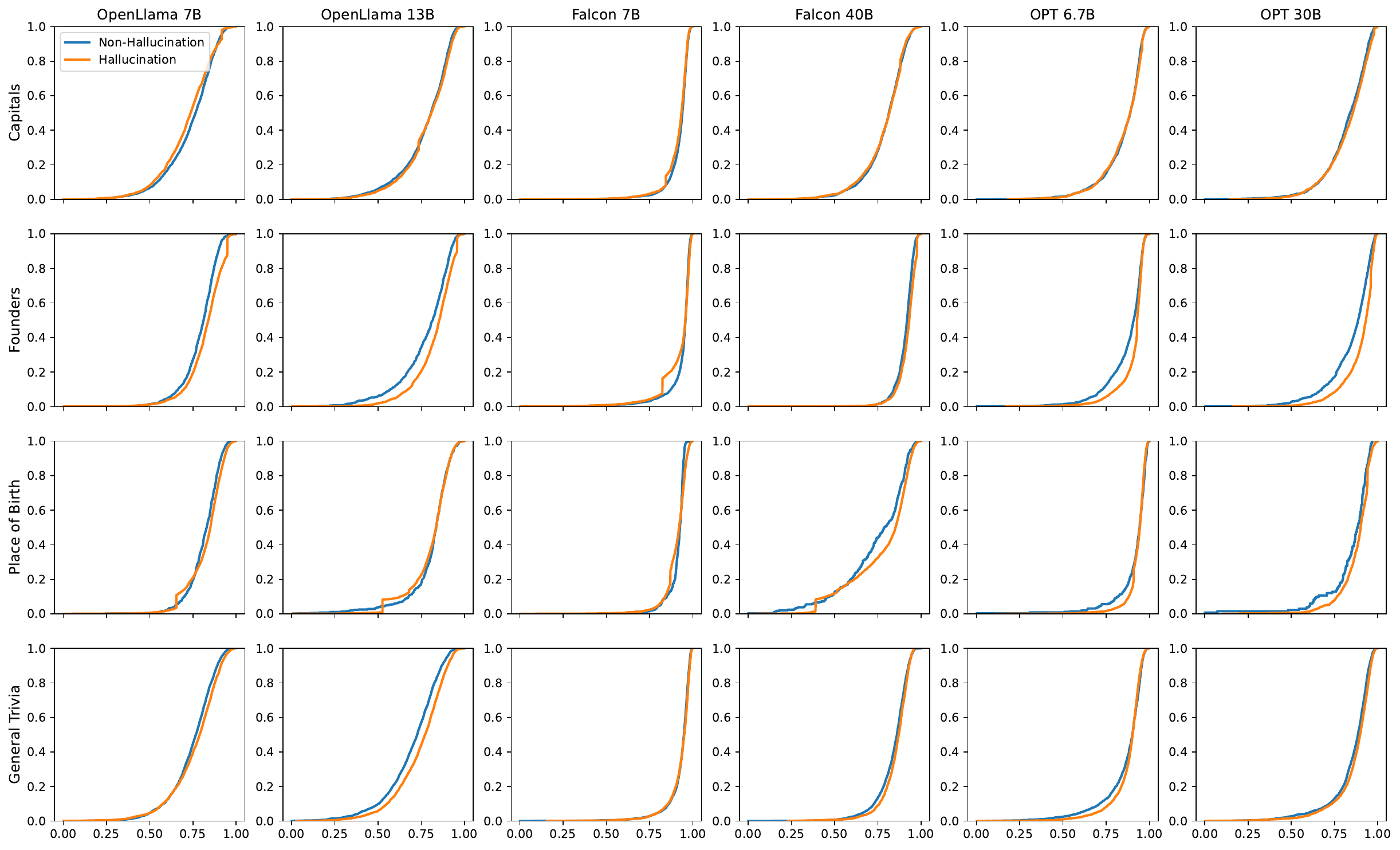}
    \caption{Cumulative distribution function of entropy in self-attention scores for all models and datasets.}
    \label{fig:attention_entropy_all}
\end{figure*}

\begin{figure*}[ht]
    \centering
    \includegraphics[width=\textwidth]{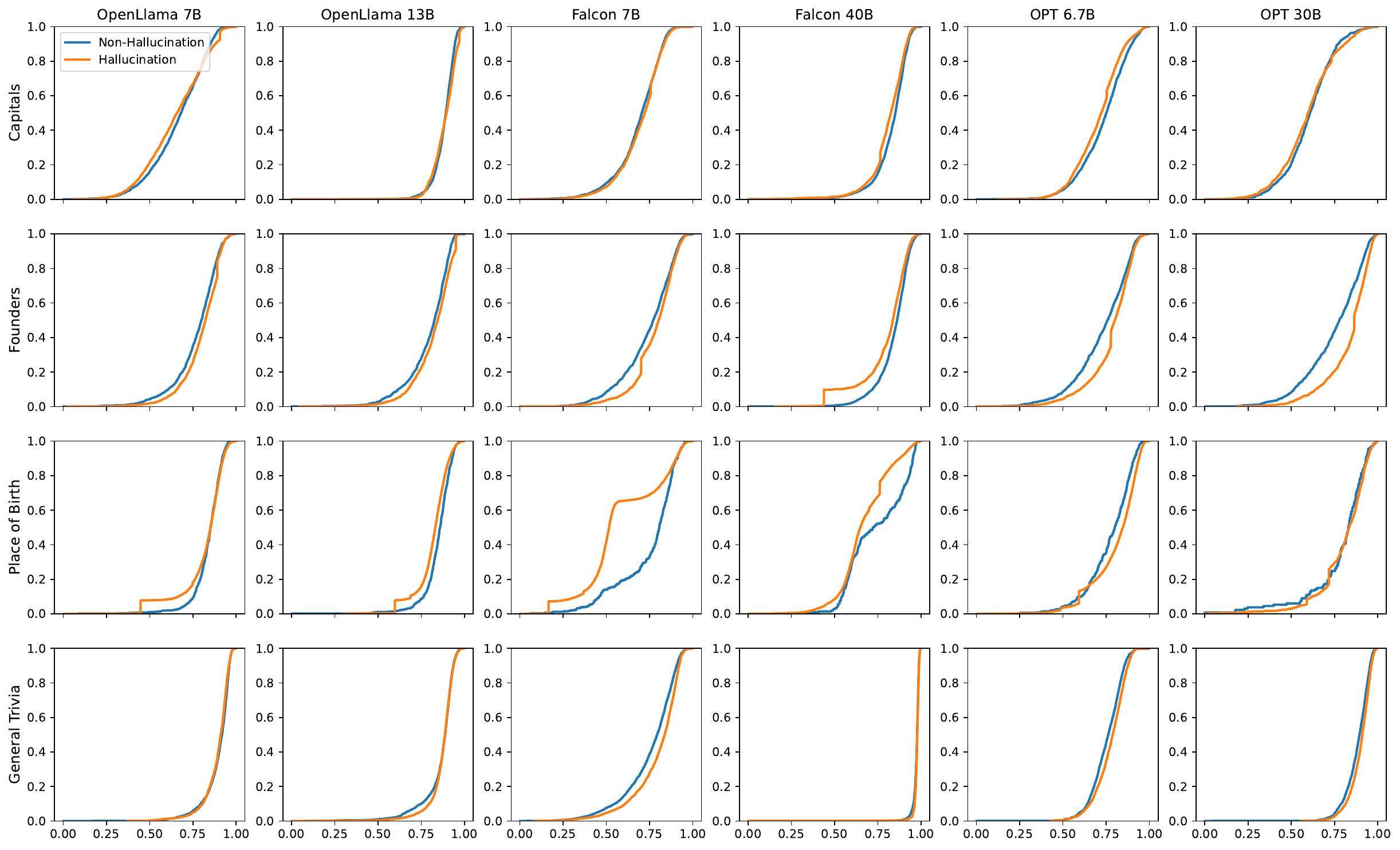}
    \caption{Cumulative distribution function of entropy in fully-connected activations for all models and datasets.}    
    \label{fig:fc_entropy_all}
\end{figure*}

\begin{figure*}[ht]
    \centering
    \includegraphics[width=\textwidth]{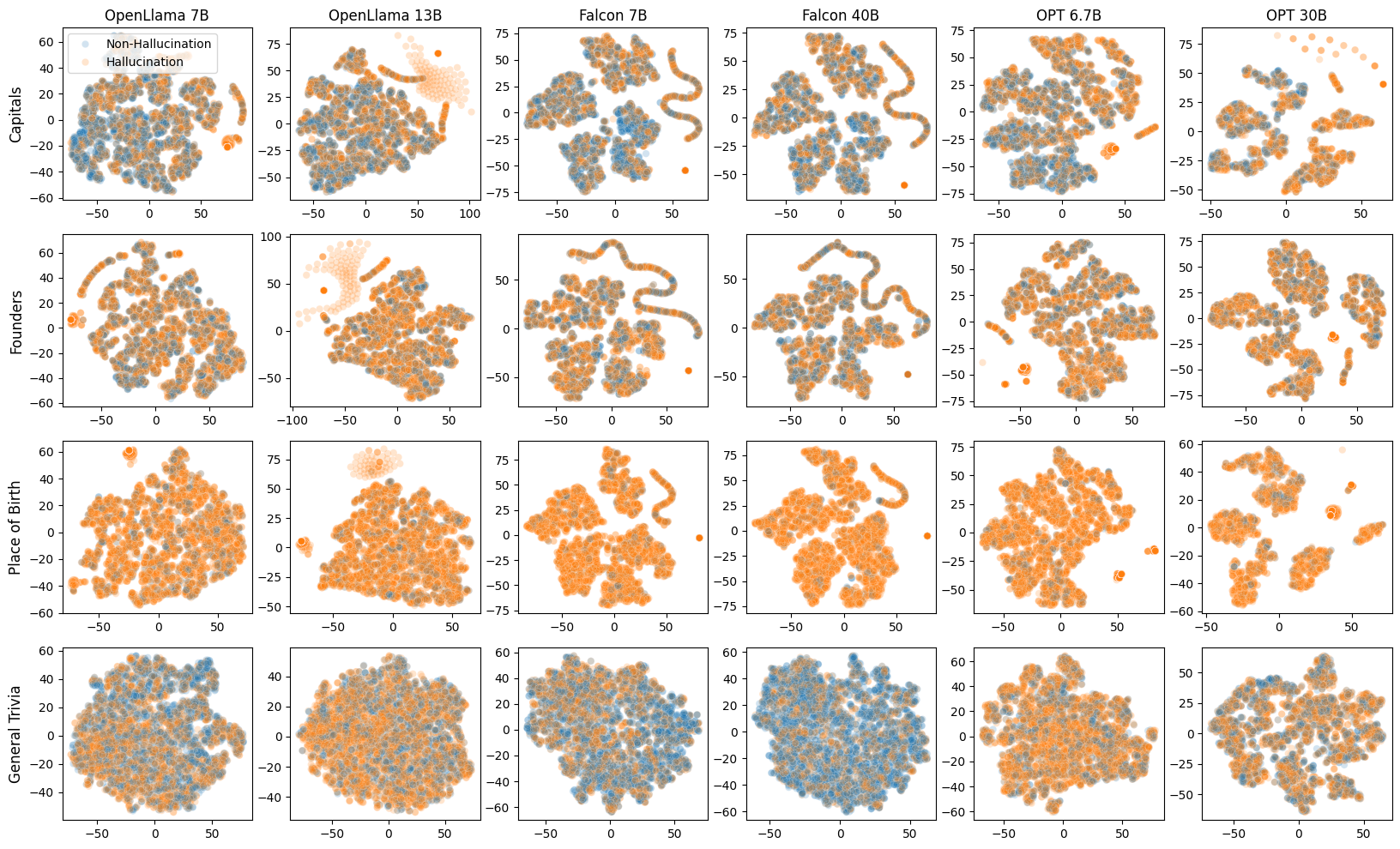}
    \caption{TSNE clustering of IG attributions for all datasets and models.}
    \label{fig:ig_tsne_all}
\end{figure*}
\begin{figure*}[ht]
    \centering
    \includegraphics[width=\textwidth]{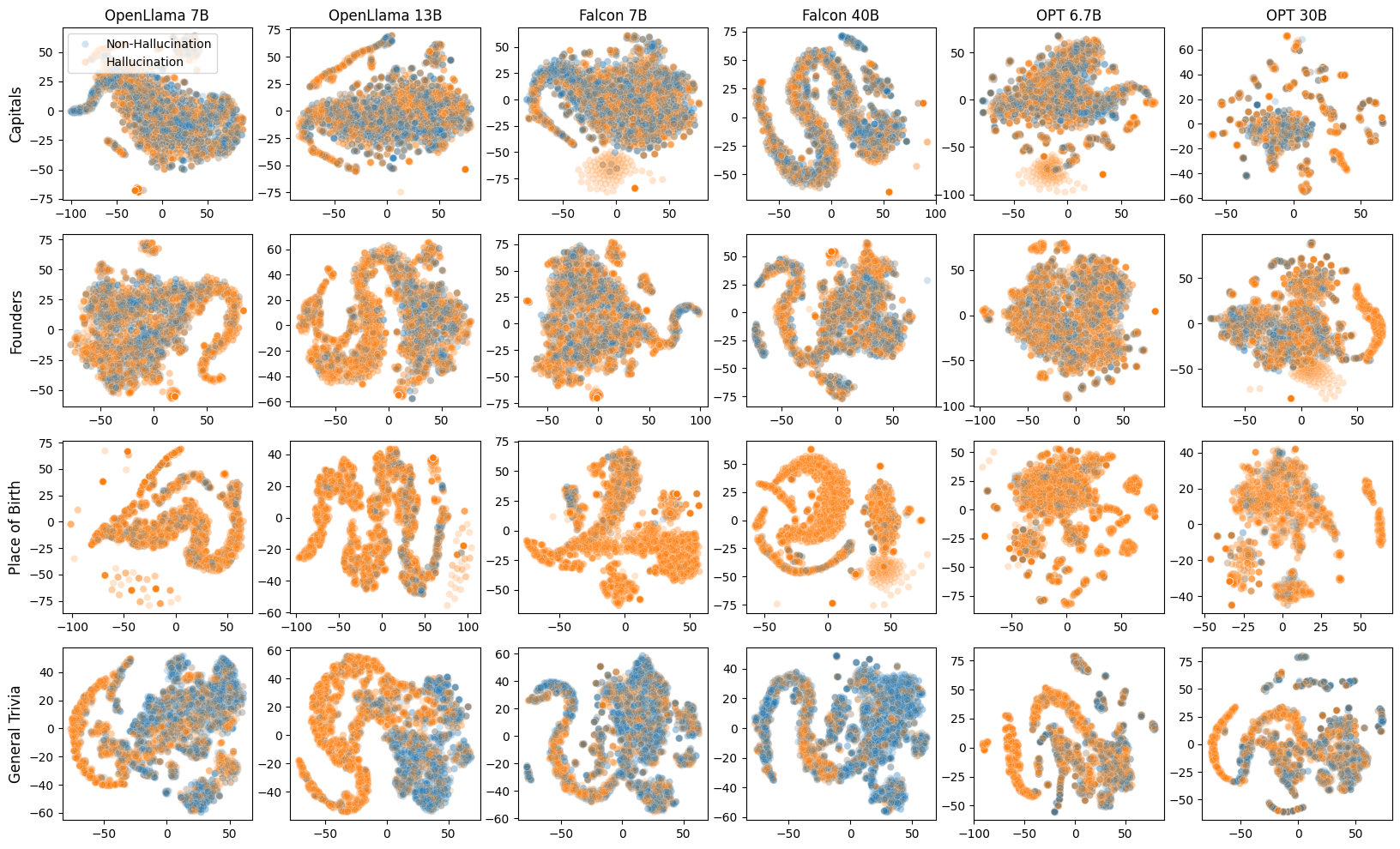}
    \caption{TSNE clustering of Softmax probabilities for all datasets and models.}
    \label{fig:softmax_tsne_all}
\end{figure*}
\begin{figure*}[ht]
    \centering
    \includegraphics[width=\textwidth]{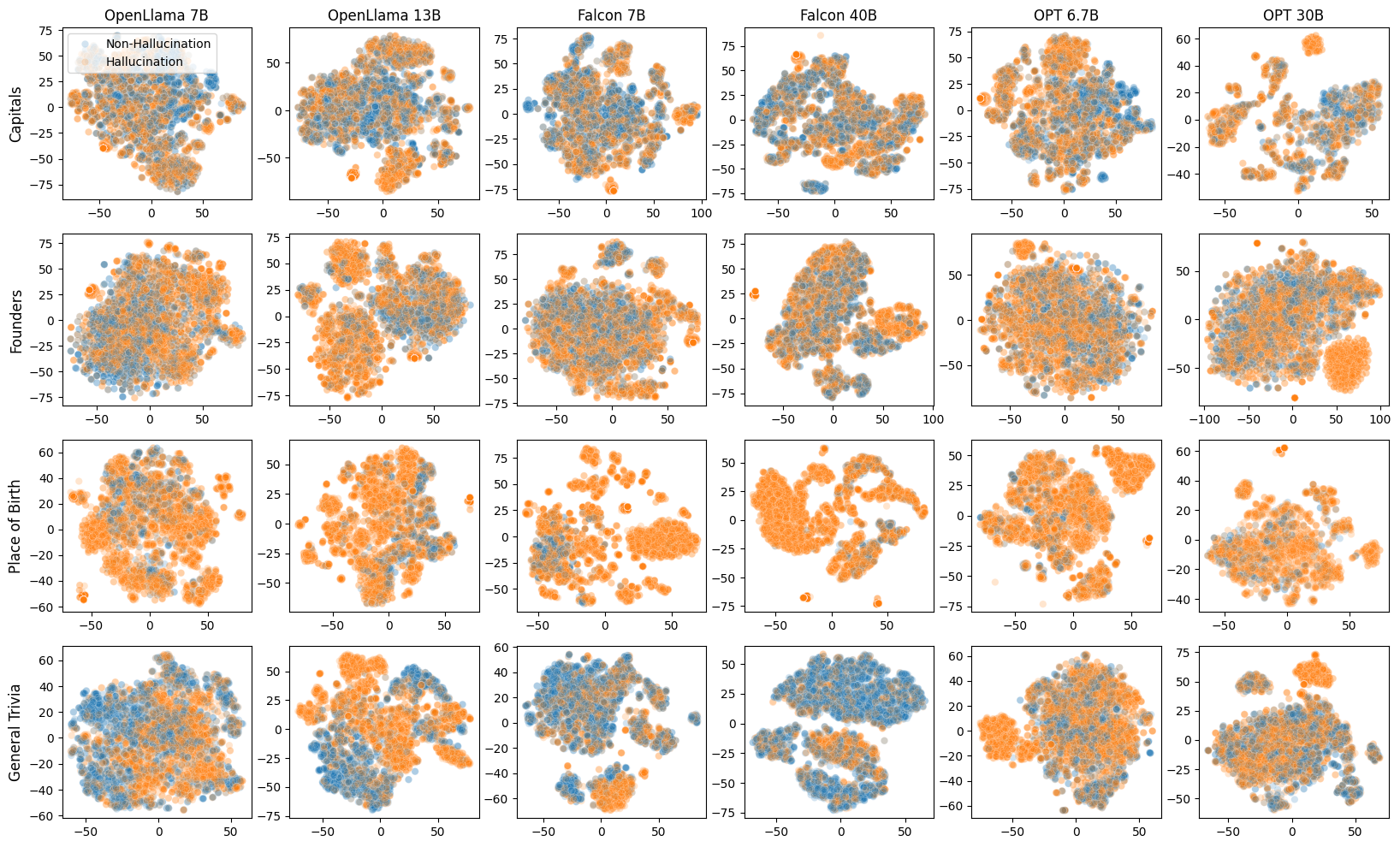}
    \caption{TSNE clustering of self-attention for all datasets and models.}
    \label{fig:attention_tsne_all}
\end{figure*}
\begin{figure*}[ht]
    \centering
    \includegraphics[width=\textwidth]{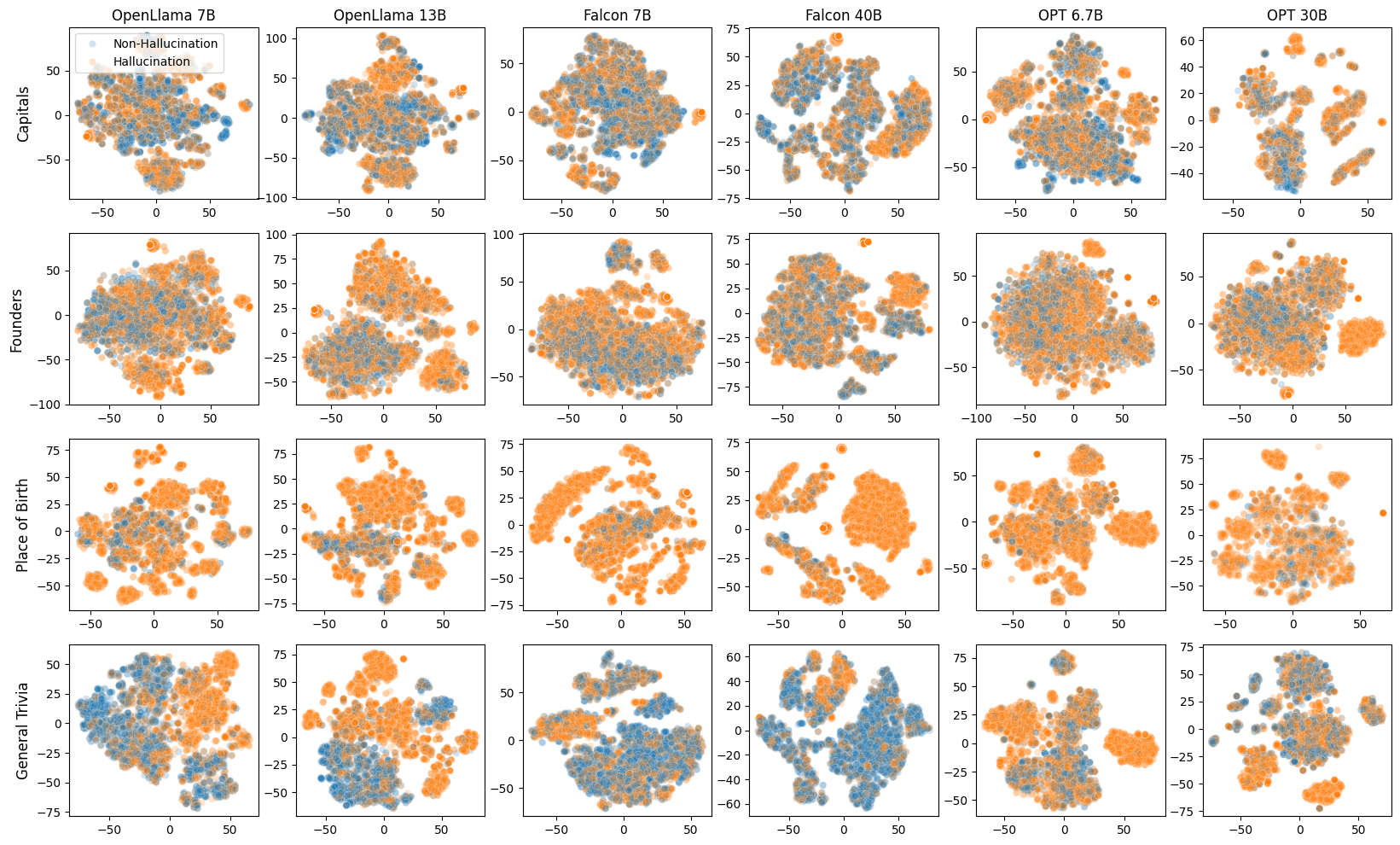}
    \caption{TSNE clustering of fully-connected activations for all datasets and models.}
    \label{fig:fc_tsne_all}
\end{figure*}

\end{document}